%% file: CoTASP.tex
\documentclass{article}

\input{math_commands}

\usepackage{microtype}
\usepackage{graphicx}
\usepackage{subfigure}
\usepackage{booktabs} 

\usepackage{hyperref}

\usepackage[accepted]{icml2023}

\usepackage{amsmath}
\usepackage{amssymb}
\usepackage{mathtools}
\usepackage{amsthm}
\usepackage{nicefrac}

\usepackage{times}
\usepackage{soul}
\usepackage{url}
\usepackage[utf8]{inputenc}
\usepackage[small]{caption}
\usepackage{graphicx}
\usepackage{booktabs}
\usepackage{amsmath,amsfonts,amssymb}

\usepackage{multirow}
\usepackage{subfigure}
\usepackage{color}
\usepackage[switch]{lineno} 
\usepackage{bbm} 
\makeatletter

\newcommand{\Rmnum}[1]{\expandafter\@slowromancap\romannumeral #1@}
\makeatother

\usepackage[capitalize,noabbrev]{cleveref}

\definecolor{mine}{RGB}{205, 232, 248}
\definecolor{mydarkred}{RGB}{139, 0, 0}

\theoremstyle{plain}

\theoremstyle{definition}

\theoremstyle{remark}

\usepackage[textsize=tiny]{todonotes}

\icmltitlerunning{Continual Task Allocation in Meta-Policy Network via Sparse Prompting}

\begin{document}

\twocolumn[
\icmltitle{Continual Task Allocation in Meta-Policy Network via Sparse Prompting}



\icmlsetsymbol{equal}{*}

\begin{icmlauthorlist}
\icmlauthor{Yijun Yang}{sustech,uts}
\icmlauthor{Tianyi Zhou}{umd}
\icmlauthor{Jing Jiang}{uts}
\icmlauthor{Guodong Long}{uts}
\icmlauthor{Yuhui Shi}{sustech}
\end{icmlauthorlist}

\icmlaffiliation{uts}{University of Technology Sydney}
\icmlaffiliation{umd}{University of Maryland, College Park}
\icmlaffiliation{sustech}{Southern University of Science and Technology}

\icmlcorrespondingauthor{Tianyi Zhou}{tianyi@umd.edu}
\icmlcorrespondingauthor{Jing Jiang}{jing.jiang@uts.edu.au}
\icmlcorrespondingauthor{Yuhui Shi}{shiyh@sustech.edu.cn}

\icmlkeywords{Machine Learning, ICML}

\vskip 0.3in
]



\printAffiliationsAndNotice{} 

\begin{abstract}
How to train a generalizable meta-policy by continually learning a sequence of tasks? It is a natural human skill yet challenging to achieve by current reinforcement learning: the agent is expected to quickly adapt to new tasks (plasticity) meanwhile retaining the common knowledge from previous tasks (stability). We address it by ``\underline{\textbf{Co}}ntinual \underline{\textbf{T}}ask \underline{\textbf{A}}llocation via \underline{\textbf{S}}parse \underline{\textbf{P}}rompting (\textbf{CoTASP})'', which learns over-complete dictionaries to produce sparse masks as prompts extracting a sub-network for each task from a meta-policy network. 
CoTASP trains a policy for each task by optimizing the prompts and the sub-network weights alternatively. 
The dictionary is then updated to align the optimized prompts with tasks' embedding, thereby capturing tasks' semantic correlations. Hence, relevant tasks share more neurons in the meta-policy network due to similar prompts while cross-task interference causing forgetting is effectively restrained. Given a meta-policy and dictionaries trained on previous tasks, new task adaptation reduces to highly efficient sparse prompting and sub-network finetuning. In experiments, CoTASP achieves a promising plasticity-stability trade-off without storing or replaying any past tasks' experiences. It outperforms existing continual and multi-task RL methods on all seen tasks, forgetting reduction, and generalization to unseen tasks. Our code is available at \url{https://github.com/stevenyangyj/CoTASP} \looseness-1
\end{abstract}

\section{Introduction}
Although reinforcement learning (RL) has demonstrated excellent performance on learning a single task, e.g., playing Go~\cite{nature/SilverHMGSDSAPL16}, robotic control~\cite{corr/ppo,nature/DegraveFBNTCEHA22}, and offline policy optimization~\cite{nips/mopo,iclr/p3}, it still suffers from catastrophic forgetting and cross-task interference when learning a stream of tasks on the fly~\cite{mccloskey1989catastrophic,icml/BengioPP20} or a curated curriculum of tasks~\cite{nips/cher,nips/co-pilot,icml/eat-c}. So it is challenging to train a meta-policy that can generalize to all learned tasks or even unseen ones with fast adaptation, which however is an inherent skill of human learning. 
This problem has been recognized as continual or lifelong RL~\cite{corr/EricEaton} and attracted growing interest in recent RL research. \looseness-1

\begin{figure}[t]
        \centering
        \includegraphics[width=0.95\linewidth]{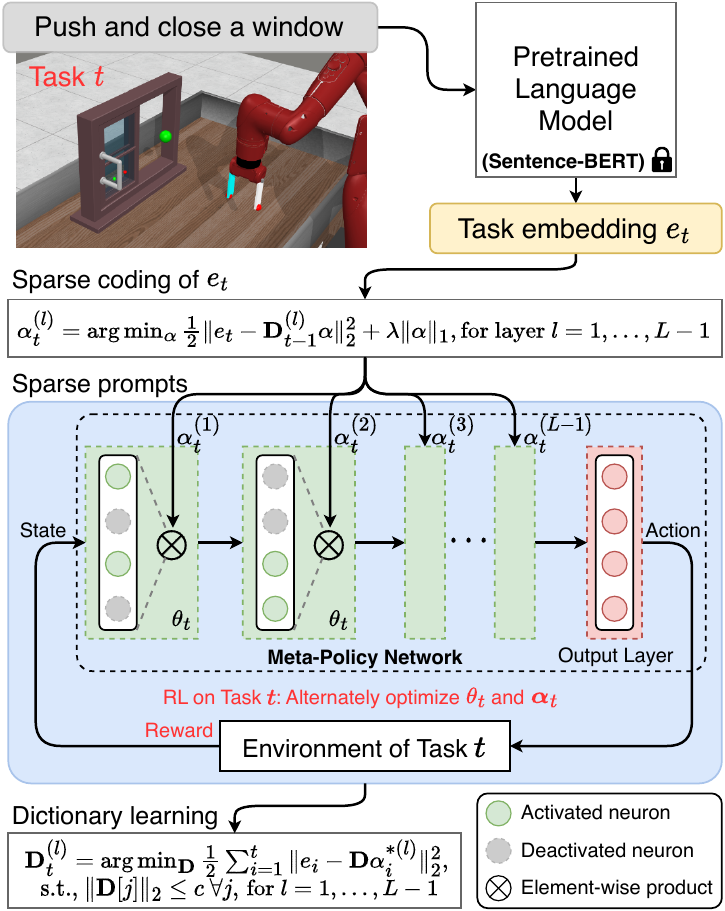}
        \caption{\footnotesize Main steps and components of CoTASP.}
        \label{fig:framework}
    \vskip -.2in
\end{figure}

A primary and long-standing challenge in continual RL is the plasticity-stability trade-off~\cite{jair/KhetarpalRRP22}: the RL policy on the one hand needs to retain and reuse the knowledge shared across different tasks in history (stability) while on the other hand can be quickly adapted to new tasks without interference from previous tasks (plasticity).
Addressing this challenge is vital to improving the efficiency of continual RL and the generalization capability of its learned policy. A meta-policy with better stability can reduce the necessity of experience replay and its memory/computation cost. Moreover, the required network size can be effectively reduced if the meta-policy can manage knowledge sharing across tasks in a more compact and efficient manner. Hence, stability can greatly improve the efficiency of continual RL when the number of tasks increases~\cite{nips/ShinLKK17,pami/LiH18a}. 
Furthermore, better plasticity indicates faster adaptation and generalization to new tasks.

In order to directly address the plasticity-stability trade-off and overcome the drawbacks of prior work, we study how to train a meta-policy network in continual RL. Then, given only a textual description of a previously learned or unseen task, a specific policy can be automatically and efficiently extracted from the meta-policy. This is in the same spirit of prompting in recent large language models~\cite{acl/prefix-tuning,corr/prompt-survey} but differs from existing methods that randomly select task policies~\cite{NEURIPS2019_rps,cvpr/MirzadehFG20} or independently optimize a policy for each task from scratch~\cite{icml/hat,icml/wsn}.
To this end, we propose to learn layer-wise dictionaries along with the meta-policy network to produce sparse prompts (i.e., binary masks) for each task, which extract a sub-network from the meta-policy to be the task-specific policy. We call this approach ``\underline{\textbf{Co}}ntinual \underline{\textbf{T}}ask \underline{\textbf{A}}llocation via \underline{\textbf{S}}parse \underline{\textbf{P}}rompting (\textbf{CoTASP})''. 

As illustrated by Fig.~\ref{fig:framework}, given each task $t$, the prompt $\alpha_t$ is generated by sparse coding of its task embedding $e_t$ under dictionary $\mathbf{D}_t$ and used to allocate a policy sub-network $\theta_t$ from the meta-policy. Then $\alpha_t$ and $\theta_t$ are optimized through RL. At the end of each task, the dictionary $\mathbf{D}_t$ is optimized~\cite{icml/odl09,colt/odl15} for all learned tasks to provide a mapping from their task embedding to their optimized prompts/sub-networks, which exploits the task correlations in both the embedding and prompt spaces. This leads to efficient usage of the meta-policy network's capacity and automatic optimization of the plasticity-stability trade-off, i.e., 
relevant tasks reuse skills by sharing more neurons (plasticity and fast adaptation) while the harmful interference between irrelevant tasks is avoided by sharing less or no neurons (stability and less forgetting).
Moreover, due to the dictionary, CoTASP does not need to store or replay any previous tasks' experiences and thus costs much less computation and memory than rehearsal-based methods~\cite{nips/clear,corr/clonex-sac}. Furthermore, the sparse prompting in CoTASP, as an efficient task adaptation method, can extract policy sub-networks for unseen tasks and thus leads to a more generalizable meta-policy. 

\begin{figure}[t]
    \vspace{-0.3em}
	\centering
	\includegraphics[width=0.75\linewidth]{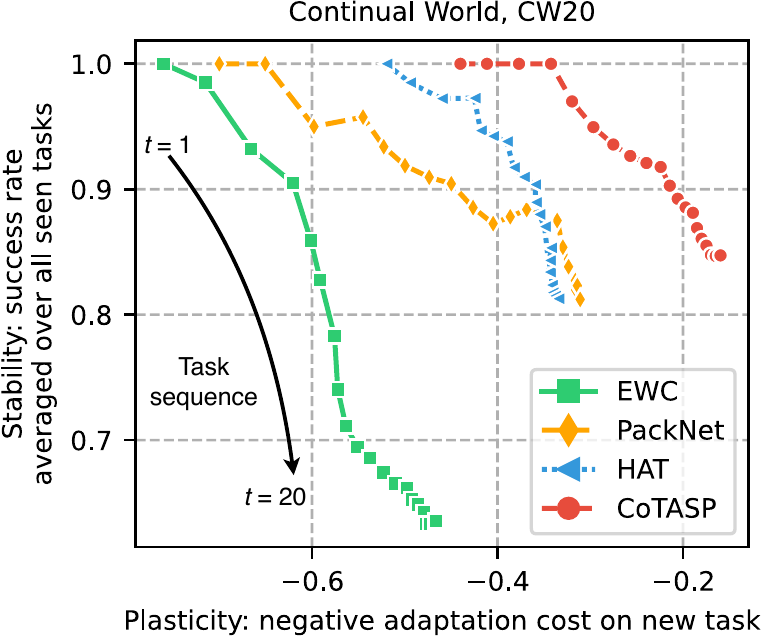}
        \vspace{-0.5em}
	\caption{\textbf{Plasticity-stability trade-off} in continual RL. Adaptation cost is the number of environment steps normalized to $(0,1]$.\looseness-1}\
        \label{fig:tradeoff}
    \vspace{-2.5em}
\end{figure}

\vspace{-0.2em}
On Continual World benchmarks~\cite{nips/continualworld}, CoTASP outperforms most baselines on all learned tasks, forgetting reduction, and generalization to unseen tasks (Table~\ref{table:main}). A thorough ablation study (Table~\ref{table:ablation}) demonstrates the importance of dictionary learning and sparse prompt optimization. Moreover, our empirical analysis shows that the dictionary converges fast (Fig.~\ref{fig:cod}) and can be generalized to future tasks, significantly reducing their adaptation cost (Fig.~\ref{fig:avg_steps}), while the learned sparse prompts capture the semantic correlations between tasks (Fig.~\ref{fig:sim_heatmap}). In comparison with state-of-the-art methods, CoTASP achieves the best plasticity-stability trade-off (Fig.~\ref{fig:tradeoff}) and highly efficient usage of model capacity (Fig.~\ref{fig:hp_effect}).

\section{Preliminaries and Related Work} \label{sec:background}
We follow the task-incremental setting adopted by prior work~\cite{jair/KhetarpalRRP22,corr/clonex-sac,nips/continualworld,nips/clear,nips/lpg-ftw,icml/p_and_c,corr/pnn}, which 
considers a sequence of tasks, each defining a Markov Decision Process (MDP) $\mathcal{M}_{t}=\langle\mathcal{S}_{t},\mathcal{A}_{t},p_{t},r_{t},\gamma\rangle$ such that $\mathcal{S}$ is the state space, $\mathcal{A}$ is the action space, $p:\mathcal{S}\times\mathcal{A}\rightarrow\Delta(\mathcal{S})$ is the transition probability where $\Delta(\mathcal{S})$ is the probability simplex over $\mathcal{S}$, $r: \mathcal{S}\times\mathcal{A}\rightarrow\mathbb{R}$ is the reward function so $r_{t}(s_{t,h},a_{t,h})$ is the immediate reward in task $t$ when taking action $a_{t,h}$ at state $s_{t,h}$, $h$ indexes the environment step, and $\gamma\in[0,1)$ is the discounted factor. Continual RL aims to achieve a policy $\pi_{\theta}$ at task $\mathcal{T}$ that performs well (with high expected return) on all seen tasks $t\leq\mathcal{T}$, with only a limited (or without) buffer of previous tasks' experiences: 
\begin{equation}
\theta^{*}=\argmax_{\theta}\sum_{t=1}^{\mathcal{T}}\mathbb{E}_{\pi_{\theta}}\left[\sum_{h=0}^{\infty}\gamma^{h}r_{t}(s_{t,h},a_{t,h})\right] \label{eq:goal_crl}
\end{equation}
Continual learning is a natural human skill that can accumulate knowledge generalizable to new tasks without forgetting the learned ones. However, 
RL agents often struggle with achieving the goal in Eq.~\ref{eq:goal_crl} due to the plasticity-stability trade-off: the policy is expected to quickly adapt to new tasks $t\geq\mathcal{T}$ (plasticity) but meanwhile to retain its performance on previous tasks $t<\mathcal{T}$ (stability).

Existing strategies for continual RL mainly focus on improving stability and reducing catastrophic forgetting. 
Rehearsal-based methods such as CLEAR~\cite{nips/clear} and P\&C~\cite{icml/p_and_c} repeatedly replay buffed experiences from previous tasks but their buffer memory and computational cost linearly grow with the number of tasks~\cite{nips/KumariW0B22}. Regularization-based methods such as EWC~\cite{pnas/ewc} and PC~\cite{icml/KaplanisSC19} alleviate forgetting without the replay buffer by adding extra regularizers when learning new tasks, which can bias the policy optimization and lead to sub-optimal solutions~\cite{icml/zhaohy}.
Finally, structure-based methods adopt different modules, i.e., sub-networks within a fixed-capacity policy network, for each task~\cite{corr/EricEaton}. We summarize two main categories of structure-based methods in the following. We also provide a more detailed discussion of related work in Appendix~\ref{appdix:related_work} and~\ref{appdix:methods}.

\textbf{Connection-level methods.~~} This category includes methods such as PackNet~\cite{cvpr/packnet}, SupSup~\cite{nips/supsup}, and WSN~\cite{icml/wsn}. 
For task $t$, the action $a_{t}$ is drawn from $a_{t}\sim\pi(s_{t};\theta\otimes \phi_{t})$ where $s_{t}$ is the state and $\phi_{t}$ is a binary mask applied to the model weights $\theta$ in an element-wise manner (i.e., $\otimes$). PackNet generates $\phi_t$ by iteratively pruning $\theta$ after the learning of each task, thereby preserving important weights for the task while leaving others for future tasks. SupSup fixes a randomly initialized network and finds the optimal $\phi_t$ for each task $t$. WSN jointly learns $\theta$ and $\phi_t$ and uses Huffman coding~\cite{huffman} to compress $\phi_t$ for a sub-linear growing size of $\{\phi_t\}^{\mathcal{T}}_{t=1}$ with increasing tasks. However, these methods usually need to 
store the task-specific masks for each task in history, leading to additional memory costs~\cite{pami/survey}. 
Moreover, their masks are seldom optimized for knowledge sharing across tasks, impeding the learned policy from being generalized to unseen tasks.

\textbf{Neuron-level methods.~~}Instead of extracting task-specific sub-networks by applying masks to model weights, the other category of methods~\cite{corr/pathnet,icml/hat,nips/ctr,ijon/spacenet} produces sub-networks by applying masks to each layer's neurons/outputs of a policy network. Compared to connection-level methods, they use layer-wise masking to achieve a more flexible and compact representation of sub-networks. But the generation of masks depends on either heuristic rules or computationally inefficient policy gradient methods~\cite{icml/nispa,icml/hat}. By contrast, CoTASP generates masks by highly efficient sparse coding (solving a relatively small lasso problem).

\vspace{0.2em}
\section{Methods}
\vspace{0.2em}
In this section, we introduce the main steps and components of CoTASP (see Fig.~\ref{fig:framework}). Specifically, Sec.~\ref{subsec:mpn} introduces the meta-policy network in the continual RL setting. Sec.~\ref{subsec:prompting} describes our proposed sparse prompting for task policy extraction. Finally, Sec.~\ref{subsec:opt_cotasp} provides the detailed optimization procedure for each component in CoTASP, including the prompt, task policy, and dictionary.

\subsection{Continual RL with a Meta-Policy Network} \label{subsec:mpn}
As discussed in Sec.~\ref{sec:background}, finetuning all weights in $\theta$ via the optimization in Eq.~\ref{eq:goal_crl} without access to past tasks leads to harmful shift on some weights important to previous tasks and catastrophic forgetting of them. 
Structure-based methods address it by allocating a sub-network for each task and 
freezing its weights once completing the learning of the task so they are immune to catastrophic forgetting. Following prior work~\cite{jmlr/dropout,corr/pathnet,icml/hat,nips/ctr,ijon/spacenet}, we represent such a sub-network by applying a binary mask to each neuron's output. Specifically, given a meta-policy network with $L$ layers, let $l\in\{1,\dots,L-1\}$ index the hidden layers of the network as a superscript, e.g., $\vy^{(l)}$ is the output vector of layer-$l$ and $\theta^{(l)}$ denotes layer-$l$'s weights. The output of the sub-network on its $(l+1)$-th layer is
\begin{equation}
    \vy^{(l+1)} = f(\phi^{(l)}_{t}\otimes\vy^{(l)}; \theta^{(l+1)}), \label{eq:subnets}
\end{equation}
where $\phi^{(l)}_{t}$ is a binary mask generated for task $t$ and applied to layer-$l$, and $f$ is a stand-in for the neural operation, e.g., a fully-connected or convolutional layer. The element-wise product operator $\otimes$ activates a portion of neurons in layer-$l$ of the meta-policy network according to $\phi^{(l)}_{t}$. These activated neurons over all layers extract a sub-network as a task-specific policy, which then interacts with the environment to collect training experiences. 
Training the sub-network avoids harmful interference with previous tasks on other neurons and meanwhile encourages their knowledge sharing on the shared neurons. 
However, allocating policies in the meta-policy network for a sequence of diverse tasks raises several challenges. For efficient usage of network capacity, each task policy should be a sparse sub-network with only a few neurons activated. Furthermore, the policy should selectively reuse neurons from previously learned policies, which can facilitate knowledge sharing between relevant tasks and reduce interference from irrelevant tasks. Inspired by prompting and in-context training for NLP~\cite{nips/RebuffiBV17,icml/adaptor-bert,acl/prefix-tuning,corr/prompt-survey}, we propose ``\textbf{Sparse Prompting}'' to address the aforementioned challenge of continual task allocation, which can automatically and efficiently extract task-specific policies from the meta-policy network. \looseness-1

\subsection{Task Policy Extraction via Sparse Prompting} \label{subsec:prompting}
In continual task allocation, the sub-network extracted for a new task is expected to reuse the knowledge learned from relevant tasks in the past and meanwhile to avoid harmful interference from irrelevant tasks. Moreover, the sub-network should be as sparse as possible for efficient usage of network capacity. Various continual RL methods~\cite{cvpr/packnet,icml/hat,ijon/spacenet,aaai/owl,icml/wsn,corr/clonex-sac} use a one-hot embedding to extract the sub-network for each learned task. They overlook semantic correlations among tasks and need delicate mechanisms to keep the sub-network sparse~\cite{pami/survey}. 

In CoTASP, we instead extract sparse sub-networks from a compact embedding of a task's textual description produced by Sentence-BERT (S-BERT)~\cite{emnlp/s-bert} via sparse coding~\cite{icml/odl09,colt/odl15}. 
In particular, we learn an over-complete dictionary $\mathbf{D}^{(l)}\in\mathbb{R}^{m\times k}$ ($m\ll k$) for each layer-$l$ of the meta-policy network, in which each column is an atom representing a neuron in the layer. 
Given a task embedding $e_{t}\in\mathbb{R}^{m}$, sparse prompting can produce a sparse prompt $\alpha_{t}^{(l)}$ for each layer-$l$ that reconstructs $e_{t}$ as a linear combination of a few neurons' representations, i.e., atoms from the dictionary. It is equal to solving the following lasso problem.
\begin{align}
    \alpha_{t}^{(l)} &= \argmin_{\alpha\in\mathbb{R}^{k}}\frac{1}{2}\|e_{t} - \mathbf{D}^{(l)}_{t-1}\alpha\|^2_2 + \lambda\|\alpha\|_1, \nonumber\\ 
    &\text{for layer~}l=1,\dots,L-1 \label{eq:sc}
\end{align}
where $\lambda$ is a regularization parameter controlling the sparsity of $\alpha$. This lasso problem can be solved by a variety of provably efficient approaches, e.g., coordinate descent~\cite{odl_cd}, fast iterative shrinkage thresholding algorithm~\cite{odl_fast}, and LARS algorithm~\cite{odl_lars}. In this paper, we adopt a Cholesky-based implementation of the LARS algorithm~\cite{icml/odl09} for its efﬁciency and stability. 
To transform $\alpha\in\mathbb{R}^{k}$ to a binary mask, we apply a step function $\sigma(\cdot)$ on $\alpha$, in which $\sigma(\alpha)=1$ if $\alpha>0$ and $0$ otherwise. 
Then we can extract a task-specific policy sub-network by applying the mask to the meta-policy network as in Eq.~\ref{eq:subnets}. 

\subsection{Meta-Policy and Dictionary Learning in CoTASP} \label{subsec:opt_cotasp}

\textbf{Alternating Optimization of Task Policy and Prompt}
By alternately optimizing the current task's policy and prompts $\alpha^{(l)}_{t}$ for $l=1,\dots,L-1$ using any off-the-shelf RL algorithm, CoTASP updates the sub-network weights associated with the task policy in the meta-policy network and the corresponding binary masks. 
However, there are two practical concerns: (1) updating the weights in $\theta$ that have already been selected by previous tasks may degrade the old tasks' performance without experience replay;  
and (2) the step function $\sigma(\cdot)$ has a zero gradient so optimizing $\alpha_{t}$ using such a gradient is infeasible. 

To address the first concern, we update the weights selectively by only allowing updates of weights that have never been allocated for any previous task. For this purpose, we accumulate the binary masks for all learned tasks by $\hat{\phi}^{(l)}_{t-1}=\vee_{i=1}^{t-1}\phi^{(l)}_{t-1}$ and update $\theta$ when learning task $t$ by
\begin{equation}
    \theta \leftarrow \theta - \eta\hat{\mathbf{g}}_{t} \nonumber
\end{equation}
\begin{equation}
    \hat{g}^{(l)}_{t} = \left\{
    \begin{aligned}
        &\left(1 - \hat{\phi}^{(l)}_{t-1}\right)g^{(l)}_{t}, &l=1 \\
        &\left(1 - \hat{\phi}^{(l-1)}_{t-1}\right)g^{(l)}_{t}, &l=L \\
        &\left[1 - \min(\hat{\phi}^{(l-1)}_{t-1},\hat{\phi}^{(l)}_{t-1})\right]g^{(l)}_{t}, &l>1
    \end{aligned}
    \right.
    \label{eq:theta_learn}
\end{equation}
where $\eta$ is the learning rate and $g^{(l)}_{t}$ denotes negative gradients of the expected return w.r.t. $\theta$ for layer-$l$ on the task $t$. In Eq.~\ref{eq:theta_learn}, we modify each weight's gradient according to the accumulated mask associated with its input and output layer. This effectively avoids overwriting the weights selected by policies of previous tasks and thus mitigates forgetting.

To address the second concern, we use the straight-through estimator (STE)~\cite{corr/ste_bengio}, i.e., $\text{clip}(\alpha,0,1)$, in the backward pass so that the $\alpha$ can be directly optimized using the same gradient descent algorithm applied to the meta-policy weights. \looseness-1

\begin{algorithm}[t]
\caption{Dictionary Learning}
\label{alg:dict_learn}
	\begin{algorithmic}[1]
	\State \textbf{input:~}$\mathbf{D}^{(l)}$ for hidden layer-$l$, $\mathbf{A}^{(l)}=[a^{(l)}_{1},\dots,a^{(l)}_{k}]\in\mathbb{R}^{k\times k}=\sum_{i=1}^{t}\alpha_{i}^{*(l)}\alpha_{i}^{*(l){\mathrm{T}}}$, $\mathbf{B}^{(l)}=[b^{(l)}_{1},\dots,b^{(l)}_{k}]\in\mathbb{R}^{m\times k}=\sum_{i=1}^{t}e_{i}\alpha_{i}^{*(l){\mathrm{T}}}$, and constant $c$
        \While{until convergence}
            \For{$j=1\text{~to~}k$}
                \State $z=\nicefrac{1}{\mathbf{A}^{(l)}_{jj}}(b^{(l)}_{j}-\mathbf{D}^{(l)}a^{(l)}_{j})+\mathbf{D}^{(l)}[j]$
                \State $\mathbf{D}^{(l)}[j]=\min\{\frac{c}{\|z\|_2},1\}z$ \Comment{\textcolor{mydarkred}{$\ell_2$-norm constraint}}
            \EndFor
        \EndWhile
        \State \textbf{output:~}updated $\mathbf{D}^{(l)}$;
	\end{algorithmic}
\end{algorithm}

\begin{algorithm}[t]
\caption{Training Procedure of CoTASP}
\label{alg:cotasp}
	\begin{algorithmic}[1]
    	\State \textbf{initialize:~}replay buffer $\mathcal{B}=\varnothing$, meta-policy network $\pi_{\theta}$ with $L$ layers, critic $Q$, dictionaries $\{\mathbf{D}^{(l)}_{0}\}_{l=1}^{L-1}$, $\mathbf{A}^{(l)}_{0}, \mathbf{B}^{(l)}_{0}, \hat{\phi}^{(l)}_{0}\leftarrow\mathbf{0}, \mathbf{0}, \mathbf{0}$, and constant $c$ for Alg.~\ref{alg:dict_learn}
	    \State \textbf{input:~}training budget $I_{\theta}$, $I_{\alpha}$, and step function $\sigma(\cdot)$
        \For{$t=1$ to $\mathcal{T}$}
            \State $e_{t}=f_{\text{S-BERT}}(\text{textual description of task~}t)$
            \State Initialize $\{\alpha_{t}^{(l)}\}_{l=1}^{L-1}$ by solving Eq.~\ref{eq:sc}
            \State Extract task-specific $\tilde{\pi}$ by Eq.~\ref{eq:subnets} with $\{\sigma(\alpha_{t}^{(l)})\}_{l=1}^{L-1}$
            \For{each iteration} \Comment{\textcolor{mydarkred}{Learning task $t$ with SAC}}
                \For{$i=1$ to $I_{\theta}$} \Comment{\textcolor{mydarkred}{Optimizing $\theta$}}
                    \State Collect $\tau=\{s_t, a_t, r_t, s_t^{\prime}\}$ with $\tilde{\pi}$
                    \State Update $\mathcal{B}$ and sample a mini-batch $\tau$
                    \State Gradient descent on $Q$
                    \State Update $\theta$ by Eq.~\ref{eq:theta_learn} with $\{\hat{\phi}^{(l)}_{t-1}\}_{l=1}^{L-1}$
                \EndFor
                \For{$i=1$ to $I_{\alpha}$} \Comment{\textcolor{mydarkred}{Optimizing $\alpha$}}
                    \State Collect $\tau=\{s_t, a_t, r_t, s_t^{\prime}\}$ with $\tilde{\pi}$
                    \State Update $\mathcal{B}$ and sample a mini-batch $\tau$
                    \State Gradient descent on $Q$
                    \State Gradient descent on $\{\alpha_{t}^{(l)}\}_{l=1}^{L-1}$ by STE
                \EndFor
            \EndFor
            \For{$l=1$ to $L-1$} \Comment{\textcolor{mydarkred}{Dictionary learning}}
                \State $\hat{\phi}^{(l)}_{t}\leftarrow\hat{\phi}^{(l)}_{t-1}\vee\sigma(\alpha_{t}^{*(l)})$
                \State $\mathbf{A}^{(l)}_{t}\leftarrow\mathbf{A}^{(l)}_{t-1}+\alpha_{t}^{*(l)}\alpha_{t}^{*(l){\mathrm{T}}}$
                \State $\mathbf{B}^{(l)}_{t}\leftarrow\mathbf{B}^{(l)}_{t-1}+e_{t}\alpha_{t}^{*(l){\mathrm{T}}}$
                \State Get updated $\mathbf{D}^{(l)}_{t}$ by Alg.~\ref{alg:dict_learn} with $\mathbf{D}^{(l)}_{t-1}$
            \EndFor
        \EndFor
        \State \textbf{output:~}$\theta^*$ and $\{\mathbf{D}^{*(l)}\}_{l=1}^{L-1}$;
	\end{algorithmic}
\end{algorithm}

\textbf{Dictionary Learning} Given the previous tasks' optimized prompts $\alpha_{i}^{*}$ and their embedding, we further update the dictionary per layer so that sparse prompting will be improved to produce the optimized prompts for each previous task., i.e., \looseness-1
\begin{align}
    \mathbf{D}^{(l)}_{t} &= \argmin_{\mathbf{D}\in\mathbb{R}^{m\times k}}\frac{1}{2}\sum_{i=1}^{t}\|e_{i} - \mathbf{D}\alpha_{i}^{*(l)}\|^2_2,\text{~s.t.,~} \|\mathbf{D}[j]\|_2\leq c \nonumber\\
    &\forall j=1,\dots,k, \text{~for layer~}l=1,\dots,L-1 \label{eq:loss_D}
\end{align}
where we constrain the $\ell_2$ norm of each atom $\mathbf{D}[j]$ to prevent the scale of $\mathbf{D}$ from growing arbitrarily large, which would lead to arbitrarily small entries in $\alpha$.

To solve the optimization with inequality constraints in Eq.~\ref{eq:loss_D}, we use block-coordinate descent with $\mathbf{D}_{t-1}^{(l)}$ as warm restart, as described in Alg.~\ref{alg:dict_learn}. Specifically, we sequentially update each atom of $\mathbf{D}_{t-1}^{(l)}$ under the constraint $\|\mathbf{D}[j]\|_2\leq c$ while fixing the rest of atoms. Since this optimization admits separable constraints to the updated atoms, convergence to a global optimum is guaranteed~\cite{bertsekas1997nonlinear,nips/sparsecoding,icml/odl09}. Moreover, $\mathbf{D}_{t-1}^{(l)}$ as a warm start for $\mathbf{D}_{t}^{(l)}$ significantly reduces the required optimization steps: we empirically found one step suffices.
The complete training procedure of CoTASP is detailed in Alg.~\ref{alg:cotasp}.

\begin{table*}[htb]
\addtolength{\tabcolsep}{-1pt}
\small
\centering
\resizebox{0.8\linewidth}{!}{
\begin{tabular}{c@{\hspace{5pt}}lr@{\hspace{-1pt}}lr@{\hspace{-1pt}}lr@{\hspace{-1pt}}l|r@{\hspace{-1pt}}lr@{\hspace{-1pt}}lr@{\hspace{-1pt}}l}
\toprule
\multicolumn{2}{l}{Benchmarks} & \multicolumn{6}{c}{CW 10} & \multicolumn{6}{c}{CW 20} \\
\midrule
\multicolumn{2}{l}{Metrics} & \multicolumn{2}{c}{$P$ ($\uparrow$)} & \multicolumn{2}{c}{$F$ ($\downarrow$)} & \multicolumn{2}{c|}{$G$ ($\downarrow$)} & \multicolumn{2}{c}{$P$ ($\uparrow$)} & \multicolumn{2}{c}{$F$ ($\downarrow$)} & \multicolumn{2}{c}{$G$ ($\downarrow$)} \\
\midrule
\multirow{5}{*}{\rotatebox[origin=c]{90}{\shortstack{Reg}}} 
&\multicolumn{1}{l}{L2} 
&\colorbox{white}{0.44}&{\color[HTML]{525252}$\pm$0.12} &\colorbox{white}{0.00}&{\color[HTML]{525252}$\pm$0.06} &\colorbox{white}{0.51}&{\color[HTML]{525252}$\pm$0.07} &\colorbox{white}{0.52}&{\color[HTML]{525252}$\pm$0.07} &\colorbox{mine}{-0.10}&{\color[HTML]{525252}$\pm$0.05} &\colorbox{white}{0.58}&{\color[HTML]{525252}$\pm$0.06}  \\
&\multicolumn{1}{l}{EWC} 
&\colorbox{white}{0.64}&{\color[HTML]{525252}$\pm$0.14} &\colorbox{white}{0.02}&{\color[HTML]{525252}$\pm$0.05} &\colorbox{white}{0.34}&{\color[HTML]{525252}$\pm$0.04} &\colorbox{white}{0.60}&{\color[HTML]{525252}$\pm$0.07} &\colorbox{white}{0.02}&{\color[HTML]{525252}$\pm$0.03} &\colorbox{white}{0.39}&{\color[HTML]{525252}$\pm$0.06} \\
&\multicolumn{1}{l}{MAS} 
&\colorbox{white}{0.60}&{\color[HTML]{525252}$\pm$0.14} &\colorbox{mine}{-0.06}&{\color[HTML]{525252}$\pm$0.04} &\colorbox{white}{0.44}&{\color[HTML]{525252}$\pm$0.07} &\colorbox{white}{0.48}&{\color[HTML]{525252}$\pm$0.06} &\colorbox{white}{0.02}&{\color[HTML]{525252}$\pm$0.02} &\colorbox{white}{0.49}&{\color[HTML]{525252}$\pm$0.03}  \\
&\multicolumn{1}{l}{VCL} 
&\colorbox{white}{0.48}&{\color[HTML]{525252}$\pm$0.10} &\colorbox{white}{-0.02}&{\color[HTML]{525252}$\pm$0.06} &\colorbox{white}{0.43}&{\color[HTML]{525252}$\pm$0.06} &\colorbox{white}{0.50}&{\color[HTML]{525252}$\pm$0.11} &\colorbox{white}{-0.04}&{\color[HTML]{525252}$\pm$0.08} &\colorbox{white}{0.52}&{\color[HTML]{525252}$\pm$0.06}  \\
&\multicolumn{1}{l}{Finetuning} 
&\colorbox{white}{0.12}&{\color[HTML]{525252}$\pm$0.04} &\colorbox{white}{0.70}&{\color[HTML]{525252}$\pm$0.04} &\colorbox{white}{0.25}&{\color[HTML]{525252}$\pm$0.06} &\colorbox{white}{0.05}&{\color[HTML]{525252}$\pm$0.00} &\colorbox{white}{0.72}&{\color[HTML]{525252}$\pm$0.03} &\colorbox{white}{0.30}&{\color[HTML]{525252}$\pm$0.05} \\
\midrule
\multirow{3}{*}{\rotatebox[origin=c]{90}{\shortstack{Struc}}} 
&\multicolumn{1}{l}{PackNet} 
&\colorbox{white}{0.80}&{\color[HTML]{525252}$\pm$0.09} &\colorbox{white}{0.00}&{\color[HTML]{525252}$\pm$0.00} &\colorbox{white}{0.28}&{\color[HTML]{525252}$\pm$0.07} &\colorbox{white}{0.78}&{\color[HTML]{525252}$\pm$0.07} &\colorbox{white}{0.00}&{\color[HTML]{525252}$\pm$0.00} &\colorbox{white}{0.32}&{\color[HTML]{525252}$\pm$0.04}  \\
&\multicolumn{1}{l}{HAT} 
&\colorbox{white}{0.68}&{\color[HTML]{525252}$\pm$0.12} 
&\colorbox{white}{0.00}&{\color[HTML]{525252}$\pm$0.00} 
&\colorbox{white}{0.44}&{\color[HTML]{525252}$\pm$0.07} 
&\colorbox{white}{0.67}&{\color[HTML]{525252}$\pm$0.08} 
&\colorbox{white}{0.00}&{\color[HTML]{525252}$\pm$0.00} 
&\colorbox{white}{0.46}&{\color[HTML]{525252}$\pm$0.04}  \\
&\multicolumn{1}{l}{TaDeLL} 
&\colorbox{white}{0.75}&{\color[HTML]{525252}$\pm$0.04} 
&\colorbox{white}{0.00}&{\color[HTML]{525252}$\pm$0.00} 
&\colorbox{white}{0.68}&{\color[HTML]{525252}$\pm$0.01} 
&\colorbox{white}{0.66}&{\color[HTML]{525252}$\pm$0.03} 
&\colorbox{white}{0.01}&{\color[HTML]{525252}$\pm$0.02} 
&\colorbox{white}{0.67}&{\color[HTML]{525252}$\pm$0.01}  \\
\midrule
\multirow{3}{*}{\rotatebox[origin=c]{90}{\shortstack{Reh}}} 
&\multicolumn{1}{l}{Reservoir} 
&\colorbox{white}{0.32}&{\color[HTML]{525252}$\pm$0.12} &\colorbox{white}{0.04}&{\color[HTML]{525252}$\pm$0.05} &\colorbox{white}{0.79}&{\color[HTML]{525252}$\pm$0.02} &\colorbox{white}{0.08}&{\color[HTML]{525252}$\pm$0.09} &\colorbox{white}{0.14}&{\color[HTML]{525252}$\pm$0.05} &\colorbox{white}{0.87}&{\color[HTML]{525252}$\pm$0.01}  \\
&\multicolumn{1}{l}{A-GEM} 
&\colorbox{white}{0.14}&{\color[HTML]{525252}$\pm$0.05} &\colorbox{white}{0.68}&{\color[HTML]{525252}$\pm$0.04} &\colorbox{mine}{0.23}&{\color[HTML]{525252}$\pm$0.02} &\colorbox{white}{0.08}&{\color[HTML]{525252}$\pm$0.02} &\colorbox{white}{0.72}&{\color[HTML]{525252}$\pm$0.07} &\colorbox{white}{0.29}&{\color[HTML]{525252}$\pm$0.04}  \\
&\multicolumn{1}{l}{ClonEx-SAC*} 
&\multicolumn{2}{c}{\colorbox{white}{0.86}} &\multicolumn{2}{c}{\colorbox{white}{0.02}} &\colorbox{white}{}&{\color[HTML]{525252}$-$} &\multicolumn{2}{c}{\colorbox{white}{0.87}} &\multicolumn{2}{c}{\colorbox{white}{0.02}} &\colorbox{white}{}&{\color[HTML]{525252}$-$}  \\
\midrule
\multirow{2}{*}{\rotatebox[origin=c]{90}{\shortstack{MT}}} 
&\multicolumn{1}{l}{MTL} 
&\colorbox{white}{0.52}&{\color[HTML]{525252}$\pm$0.10} &\colorbox{white}{}&{\color[HTML]{525252}$-$} &\colorbox{white}{}&{\color[HTML]{525252}$-$} &\colorbox{white}{0.50}&{\color[HTML]{525252}$\pm$0.11} &\colorbox{white}{}&{\color[HTML]{525252}$-$} &\colorbox{white}{}&{\color[HTML]{525252}$-$}  \\
&\multicolumn{1}{l}{MTL+PopArt} 
&\colorbox{white}{0.70}&{\color[HTML]{525252}$\pm$0.14} &\colorbox{white}{}&{\color[HTML]{525252}$-$} &\colorbox{white}{}&{\color[HTML]{525252}$-$} &\colorbox{white}{0.66}&{\color[HTML]{525252}$\pm$0.17} &\colorbox{white}{}&{\color[HTML]{525252}$-$} &\colorbox{white}{}&{\color[HTML]{525252}$-$}  \\
\midrule
\multirow{1}{*}{\rotatebox[origin=c]{90}{\shortstack{}}} 
&\multicolumn{1}{l}{CoTASP (ours)} 
&\colorbox{mine}{0.92}&{\color[HTML]{525252}$\pm$0.04} &\colorbox{white}{0.00}&{\color[HTML]{525252}$\pm$0.00} &\colorbox{white}{0.24}&{\color[HTML]{525252}$\pm$0.03} &\colorbox{mine}{0.88}&{\color[HTML]{525252}$\pm$0.02} &\colorbox{white}{0.00}&{\color[HTML]{525252}$\pm$0.00} &\colorbox{mine}{0.27}&{\color[HTML]{525252}$\pm$0.03} \\
\bottomrule
\end{tabular}}
\caption{\textbf{Evaluation (mean$\pm$std of 3 metrics over 5 random seeds) on Continual World.} $*$-reported in previous work. Reg = Regularization-based, Struc = Structure-based, Reh = Rehearsal-based, MT = Multi-task, $P$ = Average Performance, $F$ = Forgetting, $G$ = Generalization. A detailed description of baselines and metrics can be found in Sec.~\ref{subsec:exp_setup}. The best result for each metric is highlighted.}
\label{table:main}
\end{table*}

\begin{figure*}[htb]
    \centering
    \includegraphics[width=\linewidth]{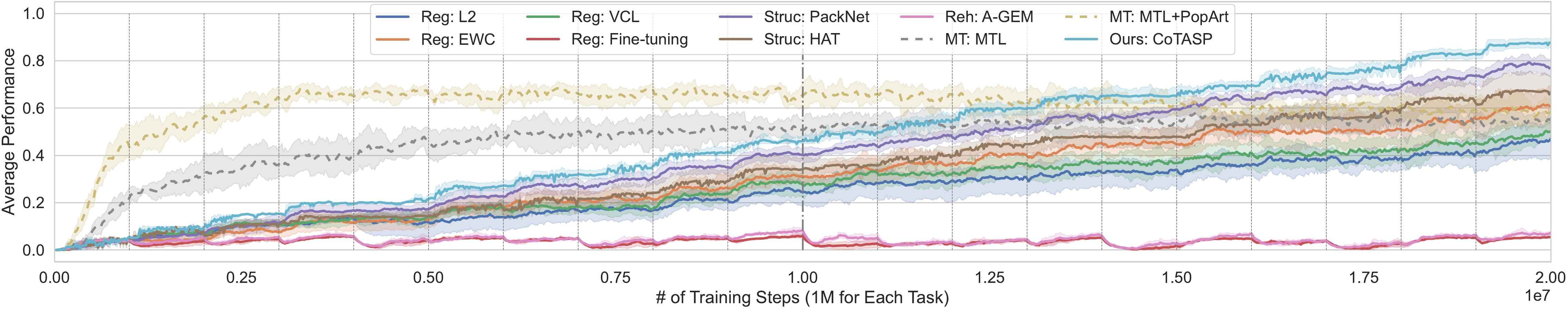}
    \caption{\textbf{Performance (mean$\pm$std over 5 random seeds) of all methods on CW20 sequence.} 
    CoTASP outperforms all the continual RL methods and all the multi-task RL baselines. We separately plot the curve per method in Fig.~\ref{fig:separate} of Appendix.}
    \label{fig:main}
\end{figure*}

\section{Experiments}\label{sec:experiment}

\subsection{Experimental Setups}\label{subsec:exp_setup}

\textbf{Benchmarks.~~}To evaluate CoTASP, we follow the same settings as prior work~\cite{corr/clonex-sac} and perform thorough experiments. Specifically, we primarily use CW10, a benchmark in the Continual World (CW)~\cite{nips/continualworld}, which consists of 10 representative manipulation tasks from MetaWorld~\cite{corl/metaworld}. To make the benchmark more challenging, we rank these tasks according to a pre-computed transfer matrix so that there is a high variation of forward transfer both in the whole sequence and locally. We also use CW20, which repeats CW10 twice, to measure the transferability of the learned policy when encountering the same task. For a fair comparison between different tasks, the number of environment interaction steps is limited to 1M per task. \looseness-1

\textbf{Evaluation metrics.~~}Following a widely-used evaluation protocol in continual learning literature~\cite{nips/gem,nips/clear,iclr/a-gem,nips/continualworld,corr/clonex-sac}, we adopt three metrics. (1) \textit{Average Performance} (higher is better): the average performance at time $t$ is defined as $P(t)=\frac{1}{\mathcal{T}}\sum_{i=1}^{\mathcal{T}}p_{i}(t)$ where $p_{i}(t)\in[0,1]$ denotes the success rate of task $i$ at time $t$. This is a canonical metric used in the continual learning community. (2) \textit{Forgetting} (lower is better): it measures the average degradation across all tasks at the end of learning, denoted by $F=\frac{1}{\mathcal{T}}\sum_{i=1}^{\mathcal{T}}p_{i}(i\cdot\delta)-p_{i}(\mathcal{T}\cdot\delta)$, where $\delta$ is the allowed environment steps for each task. (3) \textit{Generalization} (lower is better): it equals to the average number of steps needed to reach a success threshold across all tasks. Note that we stop the training when the success rate in two consecutive evaluations reaches the threshold (set to 0.9). Moreover, the metric is divided by $\delta$ to normalize its scale to $[0,1]$. \looseness-1

\textbf{Comparing methods.~~}We compare CoTASP with several baselines and state-of-the-art (SoTA) continual RL methods. According to~\cite{pami/survey}, these methods can be divided into three categories: regularization-based, structure-based, and rehearsal-based methods. Concretely, regularization-based methods include L2, Elastic Weight Consolidation (EWC)~\cite{pnas/ewc}, Memory-Aware Synapses (MAS)~\cite{eccv/mas}, and Variational Continual Learning (VCL)~\cite{iclr/vcl}. Structure-based methods include PackNet~\cite{cvpr/packnet}, Hard Attention to Tasks (HAT)~\cite{icml/hat}, and TaDeLL~\cite{jair/tadell}. Rehearsal-based methods include Reservoir, Average Gradient Episodic Memory (A-GEM)~\cite{iclr/a-gem}, and ClonEx-SAC~\cite{corr/clonex-sac}. For completeness, we also include a naive sequential training method (i.e., Finetuning) and representative multi-task RL baselines (MTL~\cite{corl/metaworld} and MTL+PopArt~\cite{aaai/popart}), which are usually regarded as the soft upper bound a continual RL method can achieve. For a fair comparison, we refer to the Continual World repository\footnote{\url{https://github.com/awarelab/continual_world}} for implementation and hyper-parameter selection. We re-run these methods to ensure the best possible performance. In addition, we adopt author-reported results for ClonEx-SAC due to the lack of open-sourced implementation. An extended description and discussion of these methods are provided in Appendix~\ref{appdix:methods}.

\textbf{Training details.~~}In order to ensure the reliability and comparability of our experiments, we follow the training details described in~\cite{nips/continualworld,corr/clonex-sac} and implement all of the baseline methods based on Soft Actor-Critic (SAC)~\cite{icml/sac}, a SoTA off-policy actor-critic algorithm. The actor and the critic are implemented as two separate multi-layer perceptron (MLP) networks, each with 4 hidden layers of 256 neurons. For structure-based methods (PackNet, HAT) and our proposed CoTASP, a wider MLP network with 1024 neurons per layer is used as the actor. We refer to these hidden layers as the backbone and the last output layer as the head. Unlike other continual RL methods~\cite{nips/gem,iclr/a-gem,icml/hat,aaai/owl} which rely on using a separate head for each new task, CoTASP uses a single-head setting where only one head is used for all tasks. In this case, CoTASP does not require selecting the appropriate head for each task and enables the reuse of parameters between similar tasks. According to~\cite{nips/continualworld}, regularizing the critic often leads to a decline in performance. Therefore, we completely ignore the forgetting issue in the critic network and retrain it for each new task. More details on the hyperparameters used in training can be found in the Appendix~\ref{appdix:hyperparameters}.

\subsection{Main Results}

This section presents the comparison between CoTASP and ten representative continual RL methods on CW benchmarks. We focus on the \textit{stability} (retain performance on seen tasks) and the \textit{plasticity} (quickly adapt to unseen tasks) and keep the constraints on computation, memory, number of samples, and neural network architecture constant. Table~\ref{table:main} summarizes our main results on CW10 and CW20 sequences. CoTASP consistently outperforms all the compared methods across different lengths of task sequences, in terms of both average performance (measures \textit{stability}) and generalization (measures \textit{plasticity}). We observe that when the hidden-layer size is the same as other structure-based methods (PackNet and HAT), CoTASP outperforms them by a large margin, especially in the generalization metric, indicating the advantage of CoTASP in improving the adaptation to new tasks. A more detailed analysis of the reasons for CoTASP's effectiveness is presented in Sec.~\ref{subsec:ablation} and \ref{subsec:effectiveness}. Moreover, we find that most continual RL methods fail to achieve positive backward transfer (i.e., $F<0$) except for VCL, suggesting the ability to improve previous tasks' performance by learning new ones is still a signiﬁcant challenge. We leave this for future work. Finally, the results in Fig.~\ref{fig:main} show that CoTASP is the only method performing comparably to the multi-task learning baselines on the first ten tasks of CW20 sequence, and it exhibits superior performance over these baselines after learning the entire CW20 sequence. One possible explanation is that the knowledge accumulated by CoTASP's meta-policy network and dictionaries leads to improved generalization. \looseness -1

\begin{table}[t]
\addtolength{\tabcolsep}{-1pt}
\small
\centering
\resizebox{0.75\linewidth}{!}{
\begin{tabular}{l@{\hspace{0pt}}r@{\hspace{-1pt}}lr@{\hspace{-1pt}}l}
\toprule
Benchmark & \multicolumn{4}{c}{CW 20} \\
\midrule
Metrics & \multicolumn{2}{c}{$P$ ($\uparrow$)} & \multicolumn{2}{c}{$G$ ($\downarrow$)} \\
\midrule
CoTASP (ours)
&\colorbox{mine}{0.88}&{\color[HTML]{525252}$\pm$0.02}  &\colorbox{mine}{0.27}&{\color[HTML]{525252}$\pm$0.03} \\
with $\mathbf{D}$ frozen 
&\colorbox{white}{0.73}&{\color[HTML]{525252}$\pm$0.06} 
&\colorbox{white}{0.47}&{\color[HTML]{525252}$\pm$0.03} \\
with $\boldsymbol{\alpha}$ frozen
&\colorbox{white}{0.79}&{\color[HTML]{525252}$\pm$0.06} 
&\colorbox{white}{0.34}&{\color[HTML]{525252}$\pm$0.02} \\
with both frozen
&\colorbox{white}{0.62}&{\color[HTML]{525252}$\pm$0.05}  &\colorbox{white}{0.52}&{\color[HTML]{525252}$\pm$0.03} \\
lazily update $\mathbf{D}$
&\colorbox{white}{\textbf{0.85}}&{\color[HTML]{525252}$\pm$0.03}  &\colorbox{white}{\textbf{0.29}}&{\color[HTML]{525252}$\pm$0.05} \\
\midrule
EWC
&\colorbox{white}{0.60}&{\color[HTML]{525252}$\pm$0.07}  &\colorbox{white}{0.39}&{\color[HTML]{525252}$\pm$0.06} \\
PackNet
&\colorbox{white}{\textbf{0.78}}&{\color[HTML]{525252}$\pm$0.07}  &\colorbox{white}{0.32}&{\color[HTML]{525252}$\pm$0.04}  \\
A-GEM
&\colorbox{white}{0.08}&{\color[HTML]{525252}$\pm$0.02}  &\colorbox{white}{\textbf{0.29}}&{\color[HTML]{525252}$\pm$0.04}  \\
Finetuning
&\colorbox{white}{0.05}&{\color[HTML]{525252}$\pm$0.00}  &\colorbox{white}{0.30}&{\color[HTML]{525252}$\pm$0.05} \\
\bottomrule
\end{tabular}}
\caption{\textbf{Ablation study.} Performance of CoTASP variants on CW20 sequence. Please refer to Sec.~\ref{subsec:ablation} for a detailed explanation.}
\label{table:ablation}
\end{table}

\subsection{Ablation Studies} \label{subsec:ablation}

\textbf{Effectiveness of core designs.~~}To show the effectiveness of each of our components, we conduct an ablation study on four variants of CoTASP, each of which removes or changes a single design choice made in the original CoTASP. Table~\ref{table:ablation} presents the results of the ablation study on CW20 sequence, using two representative evaluation metrics. Among the four variants of CoTASP, ``$\mathbf{D}$ frozen'' replaces the learnable dictionary with a fixed, randomly initialized one; ``$\boldsymbol{\alpha}$ frozen'' removes the prompt optimization proposed in Sec.~\ref{subsec:opt_cotasp}; ``both frozen'' neither updates the dictionary nor optimizes the prompt; ``lazily update $\mathbf{D}$'' stops the dictionary learning after completing the first ten tasks of CW20 sequence. According to the results in Table~\ref{table:ablation}, we give the following conclusions: (1) The use of a fixed, randomly initialized dictionary degrades the performance of CoTASP on two evaluation metrics, highlighting the importance of the learnable dictionary in capturing semantic correlations among tasks. (2) The ``$\boldsymbol{\alpha}$ frozen'' variant performs comparably to our CoTASP but outperforms the results achieved by EWC and PackNet. This indicates that optimizing the prompt can improve CoTASP's performance but is not crucial to our appealing results. (3) The ``both frozen'' variant exhibits noticeable degradation in performance, supporting the conclusion that the combination of core designs proposed in CoTASP is essential for achieving strong results. (4) The ``lazily update $\mathbf{D}$'' variant only slightly degrades from the original CoTASP on the performance but still outperforms all baselines by a large margin, indicating that the learned dictionary has accumulated sufficient knowledge in the first ten tasks so that CoTASP can achieve competitive results without updating the dictionary for repetitive tasks.

\begin{figure}[t]
	\centering
	\includegraphics[width=0.7\linewidth]{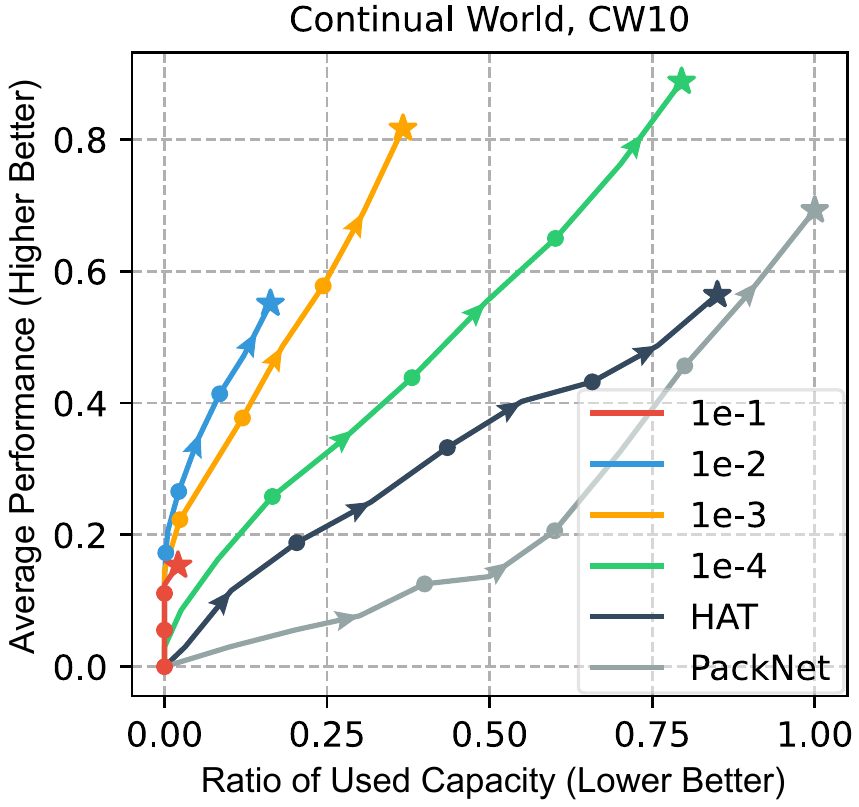}
	\caption{\textbf{Model Capacity Usage.} Comparison of CoTASP with different sparsity $\lambda$ and two baselines on the efficiency of model capacity usage, i.e., ratio of used parameters vs. performance.\looseness-1 
 }
    \label{fig:hp_effect}
\end{figure}

\textbf{Effect of key hyperparameters.~~}CoTASP introduces the sparsity parameter $\lambda$, a hyperparameter that controls the trade-off between the used network capacity and the performance of the resulting policy. A larger value of $\lambda$ results in a more sparse policy sub-network, improving the usage efficiency of the meta-policy network's capacity. But the cost is decreased performance on each task due to the loss of expressivity of the over-sparse task policy. According to the results in Fig.~\ref{fig:hp_effect}, CoTASP with $\lambda$=1e-3 or 1e-4 achieves better trade-off between performance and usage efficiency than other structure-based methods (HAT and PackNet) on CW10 sequence.\looseness-1

\begin{figure*}[!t]
    \small
        \subfigure[Results of Task Adaptation]{\includegraphics[width=0.69298\textwidth]{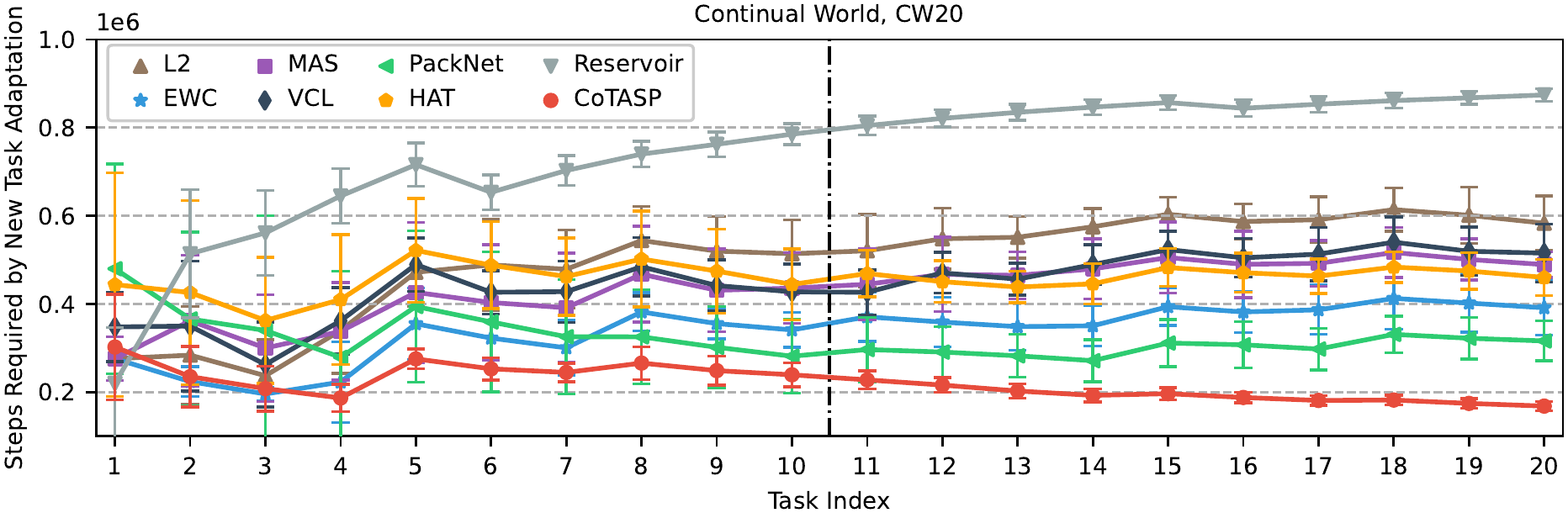}
        \label{fig:avg_steps}
        }
        \subfigure[Results of Dictionary Learning]{\includegraphics[width=0.28702\textwidth]{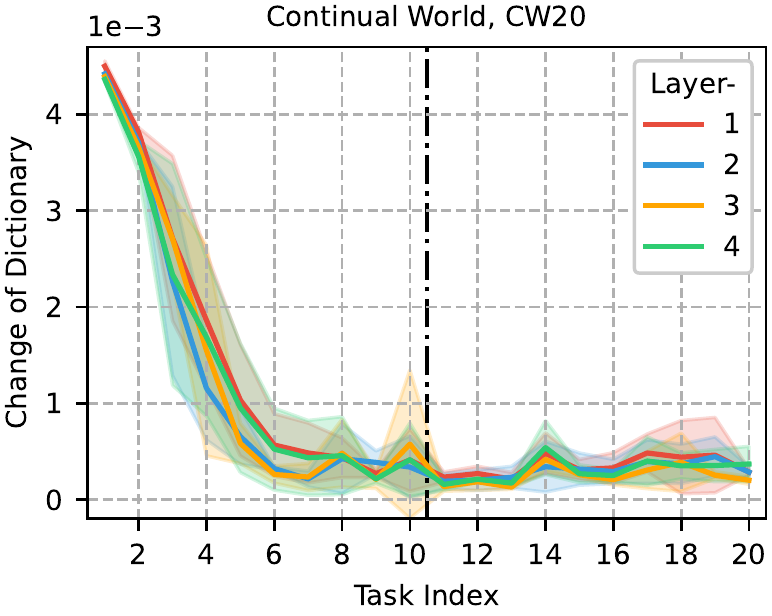}
        \label{fig:cod}
        }
    \caption{\textbf{(a):} Steps (mean$\pm$std over 5 random seeds) required by new task adaptation after learning $t$ tasks (fewer is better) on CW20 sequence. \textbf{CoTASP spends the least steps on the new task adaptation and the steps decrease for later tasks}, indicating a benefit of knowledge reuse. \textbf{(b):} The change of dictionary (averaged over 5 random seeds) of each hidden layer on CW20 sequence. The change is computed by $\nicefrac{1}{|\mathbf{D}^{(l)}_{t}|}\|\mathbf{D}^{(l)}_{t}-\mathbf{D}^{(l)}_{t-1}\|^2_2$. The black dash-dotted line splits the x-axis in two parts, the first CW10 sequence (left part) and the repeated one (right part). It shows \textbf{a fast convergence of the dictionary learning}. Please refer to Sec.~\ref{subsec:effectiveness} for a detailed discussion.}
\end{figure*}

\subsection{Why does CoTASP work? An Empirical Study} \label{subsec:effectiveness}

In this section, we answer the following questions based on the phenomena observed from our empirical results: (1) Is the learned dictionary generalizable? (2) Does the sparse prompts generated by CoTASP capture the semantic correlations between tasks?

\begin{figure}[htb]
    \centering
    \includegraphics[width=0.98\linewidth]{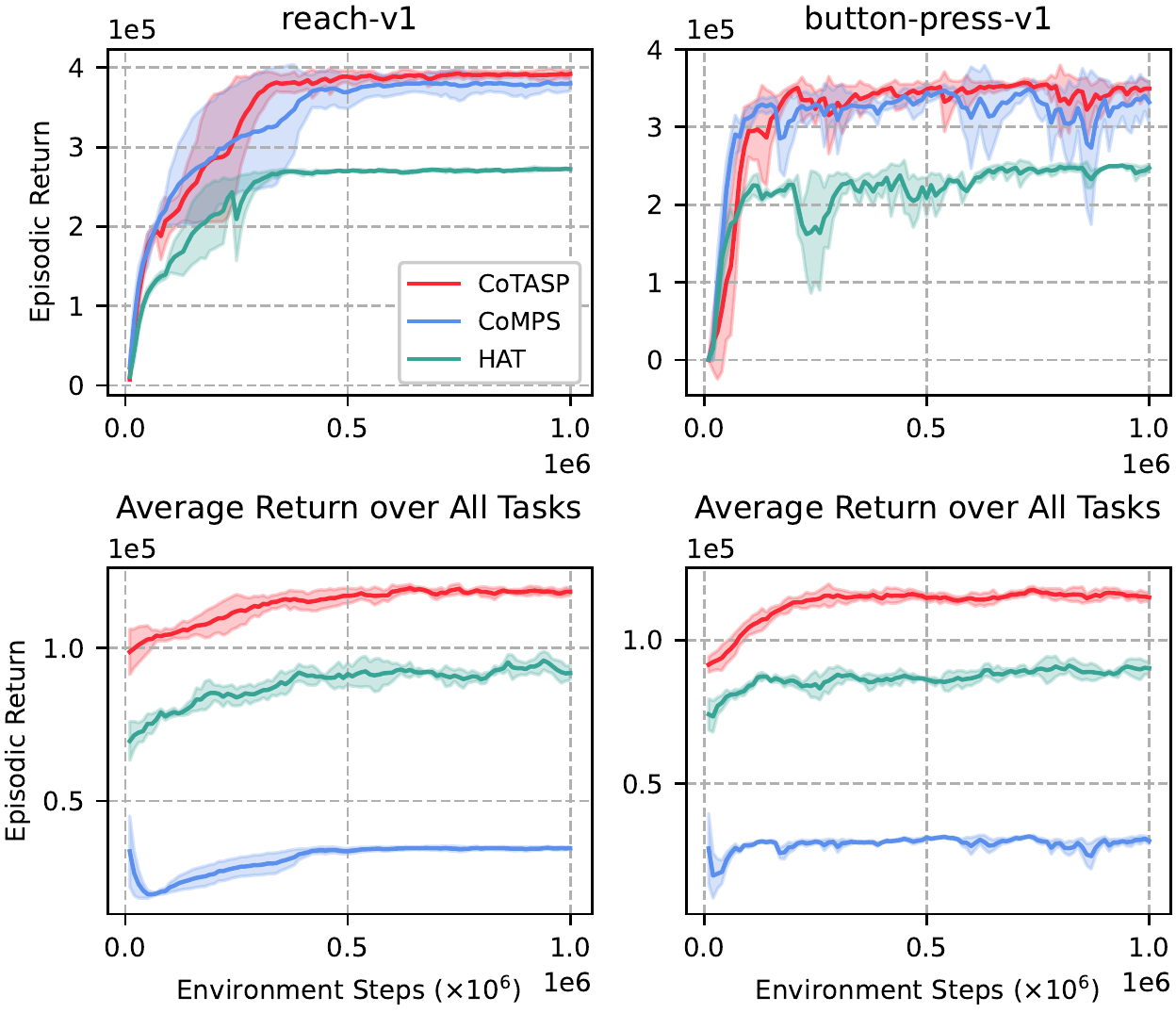}
    \caption{\textbf{Evaluation (mean$\pm$std over 6 random seeds) in Meta-World environments}~\cite{corl/metaworld}. 
    The complete results in 5 environments are reported in  Fig.~\ref{fig:ft_compare_ex} of Appendix.}
    \label{fig:ft_compare}
\end{figure}

To answer the first question, we measure the change of dictionaries, visualize their dynamics, and compare the adaptation cost of CoTASP with other continual/meta RL methods on CW20 sequence. The results in Fig~\ref{fig:cod} show that these dictionaries converge quickly with the increasing number of tasks, leading to a stationary mapping from task embedding to the optimized prompt. In next new task, CoTASP will produce a ``good'' initial prompts by sparse coding of its task embedding. This significantly reduces the number of training steps needed to reach the success threshold, as demonstrated by Fig~\ref{fig:avg_steps}. Furthermore, we compare CoTASP with a SoTA meta-RL algorithm, CoMPS~\cite{iclr/comps}, to show its superiority. Specifically, we pre-train CoTASP and CoMPS on CW10 sequence and then use the learned meta-policy as the initial policy to fine-tune unseen tasks, e.g., reach-v1 and button-press-v1. According to the results shown in Fig~\ref{fig:ft_compare}, CoTASP performs comparably to CoMPS but significantly better than the continual RL baseline (HAT) in terms of adaptation cost. However, due to the lack of mechanism against catastrophic forgetting, CoMPS adapts to the new task while its performance on previous tasks quickly degrades, resulting in worse average return across all tasks.

\begin{figure}[htb]
    \centering
    \includegraphics[width=0.95\linewidth]{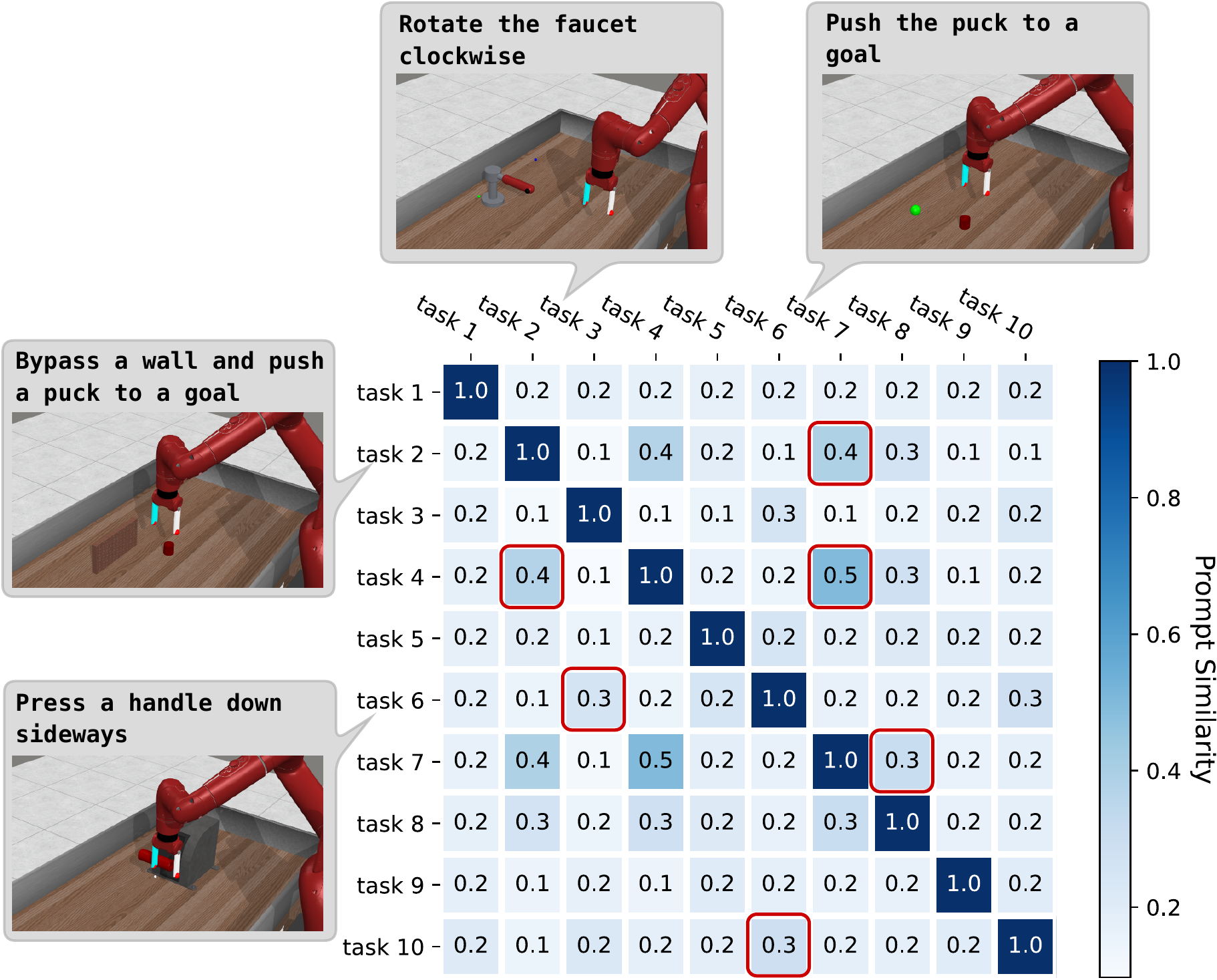}
    \caption{\textbf{Prompt (sub-network mask) similarity between two tasks} $i$ and $j$ computed by $\frac{1}{L-1}\sum_{l}\nicefrac{\|\phi^{(l)}_{i}\wedge\phi^{(l)}_{j}\|_1}{\|\phi^{(l)}_{i}\vee\phi^{(l)}_{j}\|_1}$. A layer-wise version is provided in Fig.~\ref{fig:similarity_full} of Appendix.}
    \label{fig:sim_heatmap}
\end{figure}

To answer the second question, we visualize the similarity (i.e., the overlap between two binary prompts) of the prompts generated by CoTASP between every two tasks over CW10 sequence in Fig.~\ref{fig:sim_heatmap}. The blue heatmap summarizes the similarity values averaged over all hidden layers. Specifically, the element on row $i$ and column $j$ is the averaged similarity value computed between task $i$ and task $j$. For task 2 and task 7, their task descriptions share the same manipulation primitive, i.e., \textit{pushing a puck}. Hence, the prompts generated by solving the lasso problem in Eq.~\ref{eq:sc} are highly correlated. By contrast, for task 2 and task 7 with irrelevant task descriptions, CoTASP produces different prompts, reducing cross-task interference and improving plasticity. \looseness -1

\section{Conclusions}
We propose CoTASP to address two key challenges in continual RL: (1) training a meta-policy generalizable to all seen and even unseen tasks, and (2) efficiently extracting a task policy from the meta-policy. 
CoTASP learns a dictionary to produce sparse masks (prompts) to extract each task's policy as a sub-network of the meta-policy and optimizes the sub-network via RL.
This encourages knowledge sharing/reusing among relevant tasks while reducing harmful cross-task interference that causes forgetting and poor new-task adaptation.
Without any experience replay, CoTASP achieves a significantly better plasticity-stability trade-off and more efficient network capacity allocation than baselines. Its extracted policies outperform all baselines on both previous and new tasks. 


\section*{Acknowledgements}
Yijun Yang and Yuhui Shi are partially supported by the Science and Technology Innovation Committee Foundation of Shenzhen under the Grant No. JCYJ20200109141235597 and ZDSYS201703031748284, National Science Foundation of China under grant number 61761136008, Shenzhen Peacock Plan under Grant No. KQTD2016112514355531, and Program for Guangdong Introducing Innovative and Entrepreneurial Teams under grant number 2017ZT07X386. We would like to thank ICML area chairs and anonymous reviewers for their efforts in reviewing this paper and their constructive comments!

\bibliography{CoTASP}
\bibliographystyle{icml2023}

\clearpage
\onecolumn
\appendix
\part*{Appendix}
\section{Continual World benchmark}

\begin{figure}[!htb]
    \centering
    \includegraphics[width=0.95\linewidth]{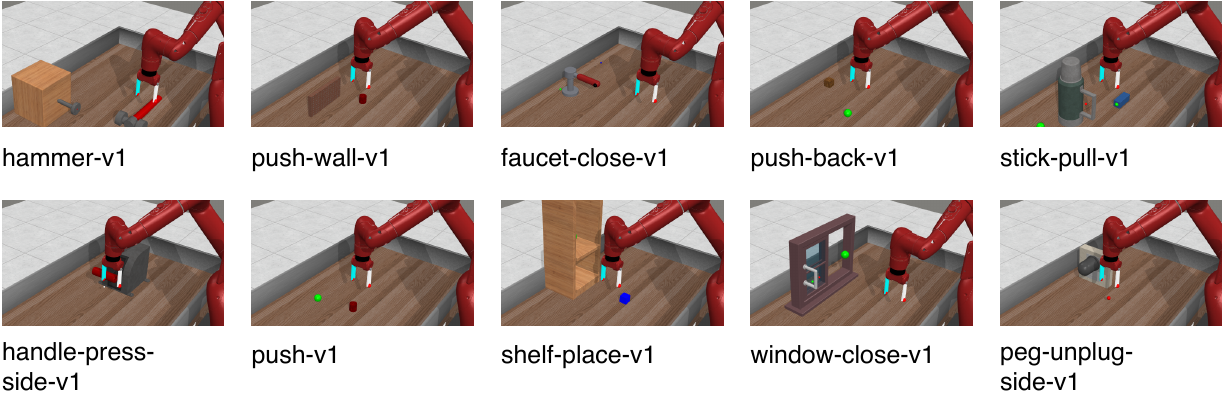}
    \caption{Continual World benchmark adopts robotic manipulation tasks developed by Meta-World~\cite{corl/metaworld}. Presented above is CW10 sequence.}
    \label{fig:bench}
\end{figure}

We visualize all of the tasks in Continual World in Fig.~\ref{fig:bench}, and provide a description of these tasks in Table~\ref{table:bench}.

\begin{table}[!htb]
\addtolength{\tabcolsep}{-1pt}
\small
\centering
\begin{tabular}{cll}
\toprule
Index & Task & Description \\
\midrule
1 & hammer-v1 & Hammer a screw on the wall. \\
2 & push-wall-v1 & Bypass a wall and push a puck to a goal. \\
3 & faucet-close-v1 & Rotate the faucet clockwise. \\
4 & push-back-v1 & Pull a puck to a goal. \\
5 & stick-pull-v1 & Grasp a stick and pull a box with the stick. \\
6 & handle-press-side-v1 & Press a handle down sideways. \\
7 & push-v1 & Push the puck to a goal. \\
8 & shelf-place-v1 & Pick and place a puck onto a shelf. \\
9 & window-close-v1 & Push and close a window. \\
10 & peg-unplug-side-v1 & Unplug a peg sideways. \\ 
\bottomrule
\end{tabular}
\caption{A list of all of the Continual-World tasks and a description of each task. Listed above is CW10 sequence. CW20 sequence contains tasks from CW10 repeated twice. Tasks are learned sequentially, and 1M environment interaction steps are allowed per task.}
\label{table:bench}
\end{table}

\section{Detailed Related Work}\label{appdix:related_work}
A promising strategy to address the stability-plasticity trade-off is using diﬀerent modules, i.e., sub-networks within a ﬁxed-capacity policy network, for each task. The modules of a new task are ﬂexible to learn (for high plasticity), while the modules of past tasks are ﬁxed (for high stability). In addition, there may exist shared modules that are trained across similar tasks, encouraging knowledge transfer. We introduce existing continual RL methods that adopt the strategy in two groups: connection-level and neuron-level methods, and also discuss their limitations.

\textbf{Connection-level methods.~~}Continusal RL methods in this category, such as PackNet~\cite{cvpr/packnet}, Piggyback~\cite{eccv/piggyback}, SupSup~\cite{nips/supsup}, BatchE~\cite{iclr/batche}, WSN~\cite{icml/wsn} and NISPA~\cite{icml/nispa}, sequentially learn and select an optimal sub-network for each task. Specifically, they alternately optimize the model weights and the binary masks of sub-networks associated with each task while attempting to select a sparse set of weights to be activated by reusing weights of the prior sub-networks. Formally, given the state $s_{t}$ from task $t$, the action $a_{t}$ is computed as $a_{t}\sim\pi(s_{t},\theta\otimes \phi_{t})$ where $\phi_{t}$ denotes the binary masks for task $t$, $\otimes$ denotes element-wise product. PackNet generates $\phi$ via iterative pruning after learning each task, which preserves important weights for current task while leaving others available for the future tasks. Piggyback learns task-specific binary masks on the weights given a pre-trained model. SupSup uses a randomly initialized, ﬁxed network and finds the optimal $\phi$ for each task. BatchE learns a shared weight matrix on the ﬁrst task and then produces only a rank-one element-wise $\phi$ for each new task. WSN jointly learns the weights and task-adaptive $\phi$ and uses Huffman coding~\cite{huffman} to compress the resulting masks for a sub-linear increase in memory with respect to the number of tasks. NISPA draws inspiration from the sparse connectivity in the human brain and proposes a heuristic mechanism to generate sparse $\phi$ for each task.

\textbf{Neuron-level methods.~~}Instead of extracting sub-networks by applying masks to every model weight as methods in the first category, another group of methods produces sub-networks for each task by applying masks to each layer's output of a policy network. Specifically, given the output $h_{l}$ of neurons of layer-$l$, the neurons to be activated are selected by $h'_{l}=h_{l}\otimes\phi_{l}$, in which $\phi_{l}$ denotes the binary masks with the same shape as $h_{l}$. These selected neurons together constitute a task-specific policy, which then interacts with the environment to collect training experiences. This category includes PathNet~\cite{corr/pathnet}, HAT~\cite{icml/hat}, CTR~\cite{nips/ctr} and SpaceNet~\cite{ijon/spacenet}. PathNet first uses Evolution Strategy~\cite{nc/evo_strategy,ppsn/des} to produce masks $\{\phi_{l}\}^{L}_{l=1}$ and then learns a sub-network according to the masks as the optimal policy. On the contrary, HAT jointly learns the policy weights and the binary masks through a gradient-based optimization. CTR borrows the idea from Adapter-BERT~\cite{icml/adaptor-bert}, which adds adapters in BERT for parameter-efﬁcient transfer learning. The key distinction between CTR and Adapter-BERT is that it employs task masks to avoid forgetting. SpaceNet obtains a sub-network by compressing the sparse connections between a selected number of neurons in each layer via the proposed sparse training. Although these methods use layer-wise masking mechanisms to achieve a more flexible and compact representation of sub-networks than connection-level methods, they have a limitation in that the generation of the masks requires either heuristic rules or computationally inefficient policy gradient methods. By contrast, our CoTASP, which also falls into this category, generates masks by encoding a accurate and compact task embedding, which is produced by LLMs, as linear combinations of a small number of atoms chosen from an over-complete dictionary. This procedure can be done efﬁciently with classical optimization tools~\cite{icml/odl09}.

Most of the aforementioned methods assume the availability of task labels, in the form of one-hot task embeddings, to identify and utilize the appropriate sub-networks during the training and inference phases. However, this assumption can lead to poor performance in real-world continual RL scenarios where tasks are diverse and prohibitive to obtain labels. CoTASP addresses this issue by learning an online dictionary that automatically generates proper masks based on the textual task descriptions, which are typically much easier to acquire from human commands~\cite{corr/cap}. Furthermore, this approach captures the semantic correlation among tasks, leading to more efficient knowledge transfer and improved generalizability. \looseness-1

\section{An Extended Description of Compared Methods} \label{appdix:methods}
We now provide a detailed description of those baseline methods compared with CoTASP. Most of them are developed in the supervised continual learning setting, and require task-specific adaption to the continual RL setting. We refer to a groundbreaking work~\cite{nips/continualworld} for the implementation of these methods. Concretely,

\textbf{Regularization-based methods.~~}This line of work reduces forgetting by restricting policy parameters important for the learned tasks. The most basic method is L2. It applies a $\ell_2$ regularization to the objective function, which keeps the parameters close to the previously optimized ones. In this method, each parameter is considered to be equally important for the previous tasks. In effect, prior work demonstrates that only a few parameters are essential for retaining performance on previous tasks. Based on this observation, EWC~\cite{pnas/ewc} adopts the Fisher information as a metric to select important parameters for previous tasks. MAS~\cite{eccv/mas} uses a parameter-wise regularizer and computes the regularization weights for each parameter by estimating their impact on the output of the policy. VCL~\cite{iclr/vcl} interprets the above methods from a Bayesian perspective and adopts variational inference to minimize the KL divergence between the posterior (current distribution of parameters) and the prior (distribution of the previously optimized parameters).

\textbf{Structure-based methods} preserve a set of parameters (i.e., sub-network), which is important for previous tasks, as a task-specific policy from a large ``super-network''. PackNet~\cite{cvpr/packnet} preserves a single sub-network for each task via iterative pruning after learning the task. HAT~\cite{icml/hat} jointly learns the policy parameters and the corresponding sub-network masks through a gradient-based optimization. TaDeLL~\cite{jair/tadell} first integrates task descriptors into continual RL and then uses coupled dictionary learning to model the inter-task relationships between the task descriptions and the task-specific policies. However, dictionary learning in TaDeLL is defined on and applied to the policy parameters instead of task embeddings in CoTASP. A linear combination of policy parameters does not hold for nonlinear neural networks~\cite{nips/lpg-ftw} and the high dimensionality of the parameters will make the dictionary learning highly inefficient. So TaDeLL is not practical (in terms of both the efficiency and effectiveness) for nonlinear policy networks. \looseness -1

\textbf{Rehearsal-based methods} repeatedly replay buffered experiences from previous tasks and use them for retraining or as constraints to alleviate catastrophic forgetting. We choose a simple reservoir strategy-based baseline, which replays all the data from the past when learning a new task, to investigate whether the naive method can increase the performance of continual RL agents. Moreover, A-GEM~\cite{iclr/a-gem} projects policy gradients from the current experiences to the closest gradients guaranteeing the average performance at previous tasks does not decrease. ClonEX-SAC~\cite{corr/clonex-sac} retains some samples from previous tasks and performs behavior cloning based on them to reduce forgetting. It achieved SoTA results on Continual World benchmarks.

\section{Hyperparameter Details} \label{appdix:hyperparameters}
We carefully tune the hyperparameters for a JAX implementation of the SAC algorithm~\cite{jax2018github, jaxrl}, and they are common for all baseline methods. Moreover, we tune the hyperparameters used for CoTASP using the ﬁnal average performance on CW10 sequence as the objective. The search space and selected hyperparameters are presented in Table~\ref{table:hyperparameters}. \looseness-1

\begin{table}[htb]
\caption{Hyperparameters of CoTASP for Continual World experiments}
\begin{center}
\begin{small}
\begin{tabular}{lll}
\toprule
\textbf{Hyperparameter} & \textbf{Search Space} & \textbf{Selected Value}  \\
\midrule
SAC Hyperparameters & & \\
\midrule
Actor hidden size & $\{256, 512, 1024, 2048\}$ & $1024$ \\
Critic hidden size & $\{128, 256, 512, 1024\}$ & $256$ \\
\# of hidden layers & $\{2, 3, 4\}$ & $4$ \\
Activation function & $\{\text{Tanh, ReLU, LeakyReLU}\}$ & $\text{LeakyReLU}$ \\
Batch size & $\{128, 256\}$ & $256$ \\
Discount factor & - & 0.99 \\
Target entropy & $\{-4.0, -2.0, 0.0\}$ & $-2.0$ \\
Target interpolation & - & $5\times10^{-3}$ \\
Replay buffer size & - & $10^{6}$ \\
Exploratory steps & - & $10^4$ \\
Optimizer & - & Adam \\
Learning rate & $\{3\times10^{-4}, 1\times10^{-3}\}$ & $3\times10^{-4}$ \\
\midrule
CoTASP-specific Hyperparameters & & \\
\midrule
Sparsity parameter $\lambda$ & $\{10^{-4}, 10^{-3}, 10^{-2}, 10^{-1}\}$ & $10^{-3}$ \\
Coding algorithm & - & $\text{LARS-Lasso}$ \\
Constant $c$ & - & $1.0$ \\
\bottomrule
\end{tabular}
\end{small}
\label{table:hyperparameters}
\end{center}
\end{table} 

\begin{figure}[!htb]
    \centering
    \includegraphics[width=0.95\linewidth]{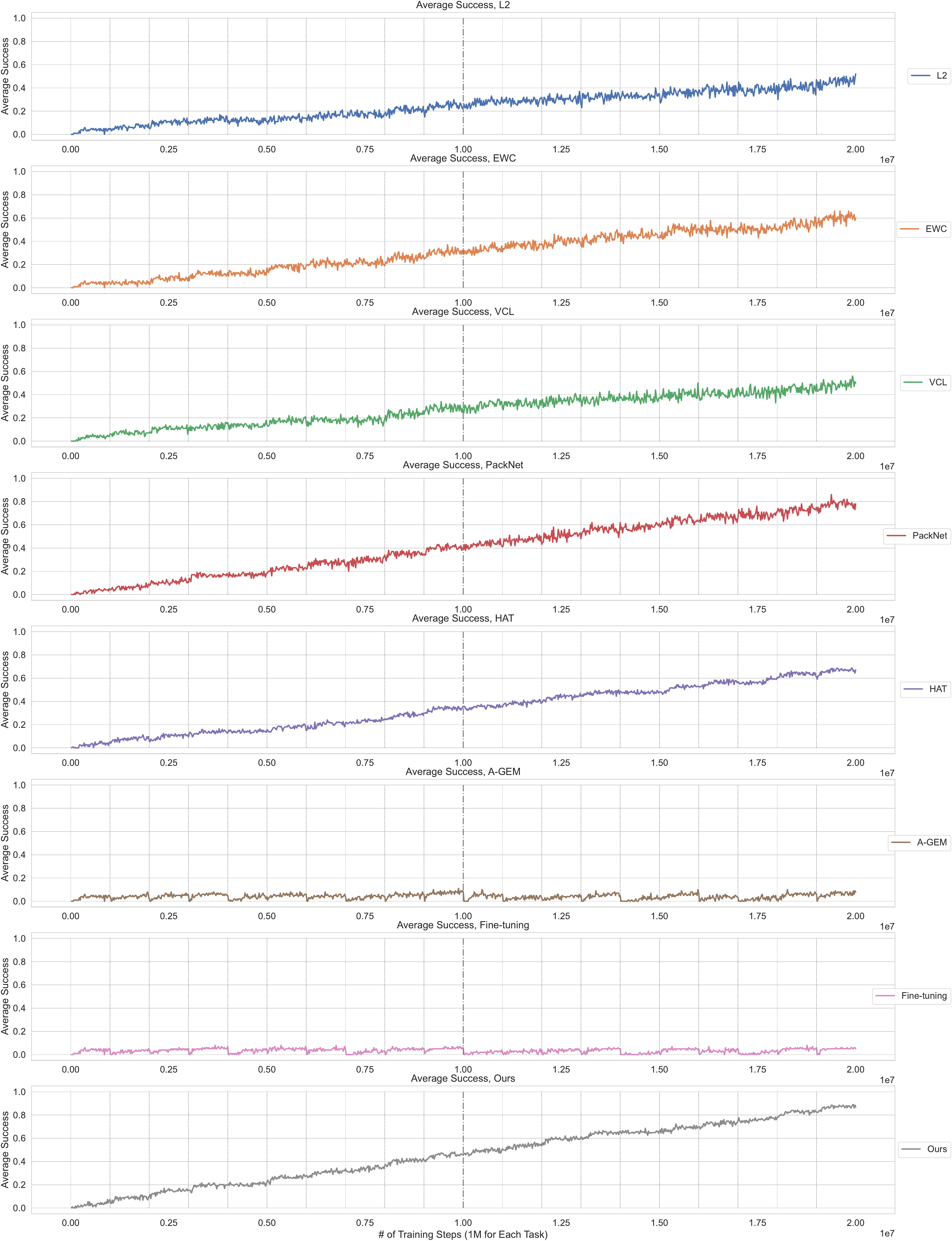}
    \caption{Average success rate over CW20 sequence for each tested method.}
    \label{fig:separate}
\end{figure}

\begin{figure}[!htb]
    \centering
    \includegraphics[width=0.95\linewidth]{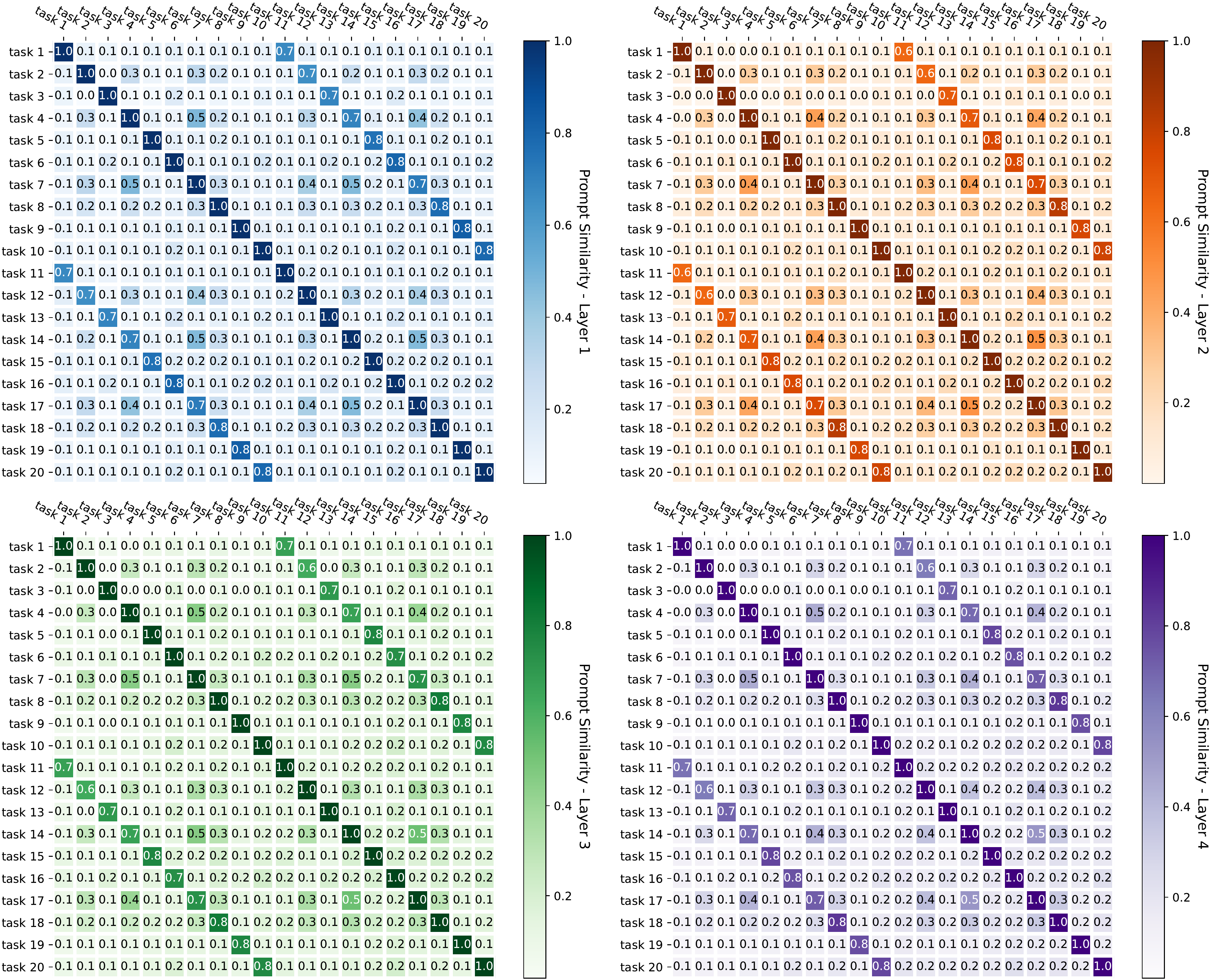}
    \caption{The similarity of each layer's prompts between every two tasks over CW20 sequence. Note that the task sequence 11-20 are repetitive CW10 sequence. We observe that the prompts generated by CoTASP are highly correlated among repetitive tasks (e.g., tasks 1 and 11), reflecting the inherent capabilities of CoTASP for knowledge transfer and task inference.}
    \label{fig:similarity_full}
\end{figure}

\begin{figure}[!htb]
    \centering
    \includegraphics[width=\linewidth]{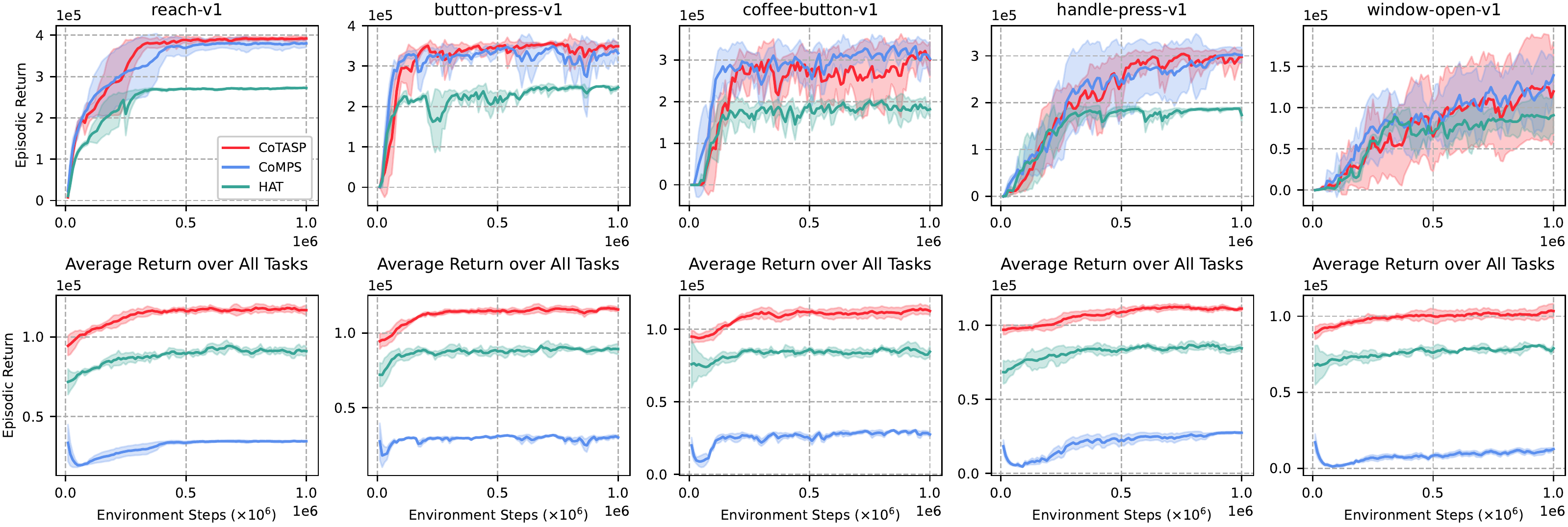}
    \caption{Results of different methods in environments from Meta-World~\cite{corl/metaworld}. All curves are the average of 6 runs with different seeds, and the shaded areas are standard errors of the mean.}
    \label{fig:ft_compare_ex}
\end{figure}

\end{document}

%% file: math_commands.tex
\usepackage{amsmath,amsfonts,bm}
\usepackage{amsthm}
\usepackage{amssymb}

\def\eqref#1{equation~\ref{#1}}

\def\1{\bm{1}}

\def\vy{{\bm{y}}}

\DeclareMathAlphabet{\mathsfit}{\encodingdefault}{\sfdefault}{m}{sl}
\SetMathAlphabet{\mathsfit}{bold}{\encodingdefault}{\sfdefault}{bx}{n}

\DeclareMathOperator*{\argmax}{arg\,max}
\DeclareMathOperator*{\argmin}{arg\,min}

%% file: CoTASP.bbl
\begin{thebibliography}{67}
\providecommand{\natexlab}[1]{#1}
\providecommand{\url}[1]{\texttt{#1}}
\expandafter\ifx\csname urlstyle\endcsname\relax
  \providecommand{\doi}[1]{doi: #1}\else
  \providecommand{\doi}{doi: \begingroup \urlstyle{rm}\Url}\fi

\bibitem[Aljundi et~al.(2018)Aljundi, Babiloni, Elhoseiny, Rohrbach, and
  Tuytelaars]{eccv/mas}
Aljundi, R., Babiloni, F., Elhoseiny, M., Rohrbach, M., and Tuytelaars, T.
\newblock Memory aware synapses: Learning what (not) to forget.
\newblock In \emph{ECCV}, 2018.

\bibitem[Ao et~al.(2021)Ao, Zhou, Long, Lu, Zhu, and Jiang]{nips/co-pilot}
Ao, S., Zhou, T., Long, G., Lu, Q., Zhu, L., and Jiang, J.
\newblock {CO-PILOT:} collaborative planning and reinforcement learning on
  sub-task curriculum.
\newblock In \emph{NeurIPS}, 2021.

\bibitem[Ao et~al.(2022)Ao, Zhou, Jiang, Long, Song, and Zhang]{icml/eat-c}
Ao, S., Zhou, T., Jiang, J., Long, G., Song, X., and Zhang, C.
\newblock {EAT-C:} environment-adversarial sub-task curriculum for efficient
  reinforcement learning.
\newblock In \emph{ICML}, 2022.

\bibitem[Arora et~al.(2015)Arora, Ge, Ma, and Moitra]{colt/odl15}
Arora, S., Ge, R., Ma, T., and Moitra, A.
\newblock Simple, efficient, and neural algorithms for sparse coding.
\newblock In \emph{COLT}, 2015.

\bibitem[Beck \& Teboulle(2009)Beck and Teboulle]{odl_fast}
Beck, A. and Teboulle, M.
\newblock A fast iterative shrinkage-thresholding algorithm for linear inverse
  problems.
\newblock \emph{SIAM journal on imaging sciences}, 2\penalty0 (1):\penalty0
  183--202, 2009.

\bibitem[Bengio et~al.(2020)Bengio, Pineau, and Precup]{icml/BengioPP20}
Bengio, E., Pineau, J., and Precup, D.
\newblock Interference and generalization in temporal difference learning.
\newblock In \emph{ICML}, 2020.

\bibitem[Bengio et~al.(2013)Bengio, L{\'{e}}onard, and
  Courville]{corr/ste_bengio}
Bengio, Y., L{\'{e}}onard, N., and Courville, A.~C.
\newblock Estimating or propagating gradients through stochastic neurons for
  conditional computation.
\newblock \emph{CoRR}, 2013.

\bibitem[Berseth et~al.(2022)Berseth, Zhang, Zhang, Finn, and
  Levine]{iclr/comps}
Berseth, G., Zhang, Z., Zhang, G., Finn, C., and Levine, S.
\newblock Comps: Continual meta policy search.
\newblock In \emph{ICLR}, 2022.

\bibitem[Bertsekas(1997)]{bertsekas1997nonlinear}
Bertsekas, D.~P.
\newblock Nonlinear programming.
\newblock \emph{Journal of the Operational Research Society}, 48\penalty0
  (3):\penalty0 334--334, 1997.

\bibitem[Beyer \& Schwefel(2002)Beyer and Schwefel]{nc/evo_strategy}
Beyer, H. and Schwefel, H.
\newblock Evolution strategies - {A} comprehensive introduction.
\newblock \emph{Nat. Comput.}, 1\penalty0 (1):\penalty0 3--52, 2002.

\bibitem[Bradbury et~al.(2018)Bradbury, Frostig, Hawkins, Johnson, Leary,
  Maclaurin, Necula, Paszke, Vander{P}las, Wanderman-{M}ilne, and
  Zhang]{jax2018github}
Bradbury, J., Frostig, R., Hawkins, P., Johnson, M.~J., Leary, C., Maclaurin,
  D., Necula, G., Paszke, A., Vander{P}las, J., Wanderman-{M}ilne, S., and
  Zhang, Q.
\newblock {JAX}: composable transformations of {P}ython+{N}um{P}y programs,
  2018.
\newblock URL \url{http://github.com/google/jax}.

\bibitem[Chaudhry et~al.(2019)Chaudhry, Ranzato, Rohrbach, and
  Elhoseiny]{iclr/a-gem}
Chaudhry, A., Ranzato, M., Rohrbach, M., and Elhoseiny, M.
\newblock Efficient lifelong learning with {A-GEM}.
\newblock In \emph{ICLR}, 2019.

\bibitem[Degrave et~al.(2022)Degrave, Felici, Buchli, Neunert, Tracey,
  Carpanese, Ewalds, Hafner, Abdolmaleki, de~Las~Casas, Donner, Fritz,
  Galperti, Huber, Keeling, Tsimpoukelli, Kay, Merle, Moret, Noury, Pesamosca,
  Pfau, Sauter, Sommariva, Coda, Duval, Fasoli, Kohli, Kavukcuoglu, Hassabis,
  and Riedmiller]{nature/DegraveFBNTCEHA22}
Degrave, J., Felici, F., Buchli, J., Neunert, M., Tracey, B.~D., Carpanese, F.,
  Ewalds, T., Hafner, R., Abdolmaleki, A., de~Las~Casas, D., Donner, C., Fritz,
  L., Galperti, C., Huber, A., Keeling, J., Tsimpoukelli, M., Kay, J., Merle,
  A., Moret, J., Noury, S., Pesamosca, F., Pfau, D., Sauter, O., Sommariva, C.,
  Coda, S., Duval, B., Fasoli, A., Kohli, P., Kavukcuoglu, K., Hassabis, D.,
  and Riedmiller, M.~A.
\newblock Magnetic control of tokamak plasmas through deep reinforcement
  learning.
\newblock \emph{Nat.}, 602\penalty0 (7897):\penalty0 414--419, 2022.

\bibitem[Duan et~al.(2022)Duan, Zhou, Shao, Yang, and Shi]{ppsn/des}
Duan, Q., Zhou, G., Shao, C., Yang, Y., and Shi, Y.
\newblock Collective learning of low-memory matrix adaptation for large-scale
  black-box optimization.
\newblock In \emph{PPSN}, 2022.

\bibitem[Efron et~al.(2004)Efron, Hastie, Johnstone, and Tibshirani]{odl_lars}
Efron, B., Hastie, T., Johnstone, I., and Tibshirani, R.
\newblock Least angle regression.
\newblock \emph{The Annals of statistics}, 32\penalty0 (2):\penalty0 407--499,
  2004.

\bibitem[Fang et~al.(2019)Fang, Zhou, Du, Han, and Zhang]{nips/cher}
Fang, M., Zhou, T., Du, Y., Han, L., and Zhang, Z.
\newblock Curriculum-guided hindsight experience replay.
\newblock In \emph{NeurIPS}, 2019.

\bibitem[Fernando et~al.(2017)Fernando, Banarse, Blundell, Zwols, Ha, Rusu,
  Pritzel, and Wierstra]{corr/pathnet}
Fernando, C., Banarse, D., Blundell, C., Zwols, Y., Ha, D., Rusu, A.~A.,
  Pritzel, A., and Wierstra, D.
\newblock Pathnet: Evolution channels gradient descent in super neural
  networks.
\newblock \emph{CoRR}, 2017.

\bibitem[Friedman et~al.(2007)Friedman, Hastie, H{\"o}fling, and
  Tibshirani]{odl_cd}
Friedman, J., Hastie, T., H{\"o}fling, H., and Tibshirani, R.
\newblock Pathwise coordinate optimization.
\newblock \emph{The annals of applied statistics}, 1\penalty0 (2):\penalty0
  302--332, 2007.

\bibitem[Gurbuz \& Dovrolis(2022)Gurbuz and Dovrolis]{icml/nispa}
Gurbuz, M.~B. and Dovrolis, C.
\newblock {NISPA:} neuro-inspired stability-plasticity adaptation for continual
  learning in sparse networks.
\newblock In \emph{ICML}, 2022.

\bibitem[Haarnoja et~al.(2018)Haarnoja, Zhou, Abbeel, and Levine]{icml/sac}
Haarnoja, T., Zhou, A., Abbeel, P., and Levine, S.
\newblock Soft actor-critic: Off-policy maximum entropy deep reinforcement
  learning with a stochastic actor.
\newblock In \emph{ICML}, 2018.

\bibitem[Hessel et~al.(2019)Hessel, Soyer, Espeholt, Czarnecki, Schmitt, and
  van Hasselt]{aaai/popart}
Hessel, M., Soyer, H., Espeholt, L., Czarnecki, W., Schmitt, S., and van
  Hasselt, H.
\newblock Multi-task deep reinforcement learning with popart.
\newblock In \emph{AAAI}, 2019.

\bibitem[Houlsby et~al.(2019)Houlsby, Giurgiu, Jastrzebski, Morrone,
  de~Laroussilhe, Gesmundo, Attariyan, and Gelly]{icml/adaptor-bert}
Houlsby, N., Giurgiu, A., Jastrzebski, S., Morrone, B., de~Laroussilhe, Q.,
  Gesmundo, A., Attariyan, M., and Gelly, S.
\newblock Parameter-efficient transfer learning for {NLP}.
\newblock In \emph{ICML}, 2019.

\bibitem[Huffman(1952)]{huffman}
Huffman, D.~A.
\newblock A method for the construction of minimum-redundancy codes.
\newblock \emph{Proceedings of the IRE}, 40\penalty0 (9):\penalty0 1098--1101,
  1952.

\bibitem[Kang et~al.(2022)Kang, Mina, Madjid, Yoon, Hasegawa{-}Johnson, Hwang,
  and Yoo]{icml/wsn}
Kang, H., Mina, R. J.~L., Madjid, S. R.~H., Yoon, J., Hasegawa{-}Johnson, M.,
  Hwang, S.~J., and Yoo, C.~D.
\newblock Forget-free continual learning with winning subnetworks.
\newblock In \emph{ICML}, 2022.

\bibitem[Kaplanis et~al.(2019)Kaplanis, Shanahan, and
  Clopath]{icml/KaplanisSC19}
Kaplanis, C., Shanahan, M., and Clopath, C.
\newblock Policy consolidation for continual reinforcement learning.
\newblock In \emph{ICML}, 2019.

\bibitem[Ke et~al.(2021)Ke, Liu, Ma, Xu, and Shu]{nips/ctr}
Ke, Z., Liu, B., Ma, N., Xu, H., and Shu, L.
\newblock Achieving forgetting prevention and knowledge transfer in continual
  learning.
\newblock In \emph{NeurIPS}, 2021.

\bibitem[Kessler et~al.(2022)Kessler, Parker{-}Holder, Ball, Zohren, and
  Roberts]{aaai/owl}
Kessler, S., Parker{-}Holder, J., Ball, P.~J., Zohren, S., and Roberts, S.~J.
\newblock Same state, different task: Continual reinforcement learning without
  interference.
\newblock In \emph{AAAI}, 2022.

\bibitem[Khetarpal et~al.(2022)Khetarpal, Riemer, Rish, and
  Precup]{jair/KhetarpalRRP22}
Khetarpal, K., Riemer, M., Rish, I., and Precup, D.
\newblock Towards continual reinforcement learning: {A} review and
  perspectives.
\newblock \emph{J. Artif. Intell. Res.}, 75:\penalty0 1401--1476, 2022.

\bibitem[Kirkpatrick et~al.(2017)Kirkpatrick, Pascanu, Rabinowitz, Veness,
  Desjardins, Rusu, Milan, Quan, Ramalho, Grabska-Barwinska, Hassabis, Clopath,
  Kumaran, and Hadsell]{pnas/ewc}
Kirkpatrick, J., Pascanu, R., Rabinowitz, N., Veness, J., Desjardins, G., Rusu,
  A.~A., Milan, K., Quan, J., Ramalho, T., Grabska-Barwinska, A., Hassabis, D.,
  Clopath, C., Kumaran, D., and Hadsell, R.
\newblock Overcoming catastrophic forgetting in neural networks.
\newblock \emph{PNAS}, 114\penalty0 (13):\penalty0 3521--3526, 2017.

\bibitem[Kostrikov(2021)]{jaxrl}
Kostrikov, I.
\newblock {JAXRL: Implementations of Reinforcement Learning algorithms in JAX},
  10 2021.
\newblock URL \url{https://github.com/ikostrikov/jaxrl}.

\bibitem[Kumari et~al.(2022)Kumari, Wang, Zhou, and Bilmes]{nips/KumariW0B22}
Kumari, L., Wang, S., Zhou, T., and Bilmes, J.~A.
\newblock Retrospective adversarial replay for continual learning.
\newblock In \emph{NeurIPS}, 2022.

\bibitem[Lange et~al.(2022)Lange, Aljundi, Masana, Parisot, Jia, Leonardis,
  Slabaugh, and Tuytelaars]{pami/survey}
Lange, M.~D., Aljundi, R., Masana, M., Parisot, S., Jia, X., Leonardis, A.,
  Slabaugh, G.~G., and Tuytelaars, T.
\newblock A continual learning survey: Defying forgetting in classification
  tasks.
\newblock \emph{{IEEE} Trans. Pattern Anal. Mach. Intell.}, 44\penalty0
  (7):\penalty0 3366--3385, 2022.

\bibitem[Lee et~al.(2006)Lee, Battle, Raina, and Ng]{nips/sparsecoding}
Lee, H., Battle, A.~J., Raina, R., and Ng, A.~Y.
\newblock Efficient sparse coding algorithms.
\newblock In \emph{NIPS}, 2006.

\bibitem[Li \& Liang(2021)Li and Liang]{acl/prefix-tuning}
Li, X.~L. and Liang, P.
\newblock Prefix-tuning: Optimizing continuous prompts for generation.
\newblock In \emph{ACL}, 2021.

\bibitem[Li \& Hoiem(2018)Li and Hoiem]{pami/LiH18a}
Li, Z. and Hoiem, D.
\newblock Learning without forgetting.
\newblock \emph{{IEEE} Trans. Pattern Anal. Mach. Intell.}, 40\penalty0
  (12):\penalty0 2935--2947, 2018.

\bibitem[Liang et~al.(2022)Liang, Huang, Xia, Xu, Hausman, Ichter, Florence,
  and Zeng]{corr/cap}
Liang, J., Huang, W., Xia, F., Xu, P., Hausman, K., Ichter, B., Florence, P.,
  and Zeng, A.
\newblock Code as policies: Language model programs for embodied control.
\newblock \emph{CoRR}, 2022.

\bibitem[Liu et~al.(2021)Liu, Yuan, Fu, Jiang, Hayashi, and
  Neubig]{corr/prompt-survey}
Liu, P., Yuan, W., Fu, J., Jiang, Z., Hayashi, H., and Neubig, G.
\newblock Pre-train, prompt, and predict: {A} systematic survey of prompting
  methods in natural language processing.
\newblock \emph{CoRR}, 2021.

\bibitem[Lopez{-}Paz \& Ranzato(2017)Lopez{-}Paz and Ranzato]{nips/gem}
Lopez{-}Paz, D. and Ranzato, M.
\newblock Gradient episodic memory for continual learning.
\newblock In \emph{NeurIPS}, 2017.

\bibitem[Mairal et~al.(2009)Mairal, Bach, Ponce, and Sapiro]{icml/odl09}
Mairal, J., Bach, F.~R., Ponce, J., and Sapiro, G.
\newblock Online dictionary learning for sparse coding.
\newblock In \emph{ICML}, 2009.

\bibitem[Mallya \& Lazebnik(2018)Mallya and Lazebnik]{cvpr/packnet}
Mallya, A. and Lazebnik, S.
\newblock Packnet: Adding multiple tasks to a single network by iterative
  pruning.
\newblock In \emph{CVPR}, 2018.

\bibitem[Mallya et~al.(2018)Mallya, Davis, and Lazebnik]{eccv/piggyback}
Mallya, A., Davis, D., and Lazebnik, S.
\newblock Piggyback: Adapting a single network to multiple tasks by learning to
  mask weights.
\newblock In \emph{ECCV}, 2018.

\bibitem[McCloskey \& Cohen(1989)McCloskey and
  Cohen]{mccloskey1989catastrophic}
McCloskey, M. and Cohen, N.~J.
\newblock Catastrophic interference in connectionist networks: The sequential
  learning problem.
\newblock In \emph{Psychology of learning and motivation}, volume~24, pp.\
  109--165. 1989.

\bibitem[Mendez \& Eaton(2022)Mendez and Eaton]{corr/EricEaton}
Mendez, J.~A. and Eaton, E.
\newblock How to reuse and compose knowledge for a lifetime of tasks: {A}
  survey on continual learning and functional composition.
\newblock \emph{CoRR}, 2022.

\bibitem[Mendez et~al.(2020)Mendez, Wang, and Eaton]{nips/lpg-ftw}
Mendez, J.~A., Wang, B., and Eaton, E.
\newblock Lifelong policy gradient learning of factored policies for faster
  training without forgetting.
\newblock In \emph{NeurIPS}, 2020.

\bibitem[Mirzadeh et~al.(2020)Mirzadeh, Farajtabar, and
  Ghasemzadeh]{cvpr/MirzadehFG20}
Mirzadeh, S., Farajtabar, M., and Ghasemzadeh, H.
\newblock Dropout as an implicit gating mechanism for continual learning.
\newblock In \emph{CVPR}, 2020.

\bibitem[Nguyen et~al.(2018)Nguyen, Li, Bui, and Turner]{iclr/vcl}
Nguyen, C.~V., Li, Y., Bui, T.~D., and Turner, R.~E.
\newblock Variational continual learning.
\newblock In \emph{ICLR}, 2018.

\bibitem[Rajasegaran et~al.(2019)Rajasegaran, Hayat, Khan, Khan, and
  Shao]{NEURIPS2019_rps}
Rajasegaran, J., Hayat, M., Khan, S.~H., Khan, F.~S., and Shao, L.
\newblock Random path selection for continual learning.
\newblock In \emph{NeurIPS}, 2019.

\bibitem[Rebuffi et~al.(2017)Rebuffi, Bilen, and Vedaldi]{nips/RebuffiBV17}
Rebuffi, S., Bilen, H., and Vedaldi, A.
\newblock Learning multiple visual domains with residual adapters.
\newblock In \emph{NeurIPS}, 2017.

\bibitem[Reimers \& Gurevych(2019)Reimers and Gurevych]{emnlp/s-bert}
Reimers, N. and Gurevych, I.
\newblock Sentence-bert: Sentence embeddings using siamese bert-networks.
\newblock In \emph{EMNLP}, 2019.

\bibitem[Rolnick et~al.(2019)Rolnick, Ahuja, Schwarz, Lillicrap, and
  Wayne]{nips/clear}
Rolnick, D., Ahuja, A., Schwarz, J., Lillicrap, T.~P., and Wayne, G.
\newblock Experience replay for continual learning.
\newblock In \emph{NeurIPS}, 2019.

\bibitem[Rostami et~al.(2020)Rostami, Isele, and Eaton]{jair/tadell}
Rostami, M., Isele, D., and Eaton, E.
\newblock Using task descriptions in lifelong machine learning for improved
  performance and zero-shot transfer.
\newblock \emph{J. Artif. Intell. Res.}, 67:\penalty0 673--704, 2020.

\bibitem[Rusu et~al.(2016)Rusu, Rabinowitz, Desjardins, Soyer, Kirkpatrick,
  Kavukcuoglu, Pascanu, and Hadsell]{corr/pnn}
Rusu, A.~A., Rabinowitz, N.~C., Desjardins, G., Soyer, H., Kirkpatrick, J.,
  Kavukcuoglu, K., Pascanu, R., and Hadsell, R.
\newblock Progressive neural networks.
\newblock \emph{CoRR}, 2016.

\bibitem[Schulman et~al.(2017)Schulman, Wolski, Dhariwal, Radford, and
  Klimov]{corr/ppo}
Schulman, J., Wolski, F., Dhariwal, P., Radford, A., and Klimov, O.
\newblock Proximal policy optimization algorithms.
\newblock \emph{CoRR}, 2017.

\bibitem[Schwarz et~al.(2018)Schwarz, Czarnecki, Luketina, Grabska{-}Barwinska,
  Teh, Pascanu, and Hadsell]{icml/p_and_c}
Schwarz, J., Czarnecki, W., Luketina, J., Grabska{-}Barwinska, A., Teh, Y.~W.,
  Pascanu, R., and Hadsell, R.
\newblock Progress {\&} compress: {A} scalable framework for continual
  learning.
\newblock In \emph{ICML}, 2018.

\bibitem[Serr{\`{a}} et~al.(2018)Serr{\`{a}}, Suris, Miron, and
  Karatzoglou]{icml/hat}
Serr{\`{a}}, J., Suris, D., Miron, M., and Karatzoglou, A.
\newblock Overcoming catastrophic forgetting with hard attention to the task.
\newblock In \emph{ICML}, 2018.

\bibitem[Shin et~al.(2017)Shin, Lee, Kim, and Kim]{nips/ShinLKK17}
Shin, H., Lee, J.~K., Kim, J., and Kim, J.
\newblock Continual learning with deep generative replay.
\newblock In \emph{NeurIPS}, 2017.

\bibitem[Silver et~al.(2016)Silver, Huang, Maddison, Guez, Sifre, van~den
  Driessche, Schrittwieser, Antonoglou, Panneershelvam, Lanctot, Dieleman,
  Grewe, Nham, Kalchbrenner, Sutskever, Lillicrap, Leach, Kavukcuoglu, Graepel,
  and Hassabis]{nature/SilverHMGSDSAPL16}
Silver, D., Huang, A., Maddison, C.~J., Guez, A., Sifre, L., van~den Driessche,
  G., Schrittwieser, J., Antonoglou, I., Panneershelvam, V., Lanctot, M.,
  Dieleman, S., Grewe, D., Nham, J., Kalchbrenner, N., Sutskever, I.,
  Lillicrap, T.~P., Leach, M., Kavukcuoglu, K., Graepel, T., and Hassabis, D.
\newblock Mastering the game of go with deep neural networks and tree search.
\newblock \emph{Nat.}, 529\penalty0 (7587):\penalty0 484--489, 2016.

\bibitem[Sokar et~al.(2021)Sokar, Mocanu, and Pechenizkiy]{ijon/spacenet}
Sokar, G., Mocanu, D.~C., and Pechenizkiy, M.
\newblock Spacenet: Make free space for continual learning.
\newblock \emph{Neurocomputing}, 439:\penalty0 1--11, 2021.

\bibitem[Srivastava et~al.(2014)Srivastava, Hinton, Krizhevsky, Sutskever, and
  Salakhutdinov]{jmlr/dropout}
Srivastava, N., Hinton, G.~E., Krizhevsky, A., Sutskever, I., and
  Salakhutdinov, R.
\newblock Dropout: a simple way to prevent neural networks from overfitting.
\newblock \emph{J. Mach. Learn. Res.}, 15\penalty0 (1):\penalty0 1929--1958,
  2014.

\bibitem[Wen et~al.(2020)Wen, Tran, and Ba]{iclr/batche}
Wen, Y., Tran, D., and Ba, J.
\newblock Batchensemble: an alternative approach to efficient ensemble and
  lifelong learning.
\newblock In \emph{ICLR}, 2020.

\bibitem[Wolczyk et~al.(2021)Wolczyk, Zajac, Pascanu, Kucinski, and
  Milos]{nips/continualworld}
Wolczyk, M., Zajac, M., Pascanu, R., Kucinski, L., and Milos, P.
\newblock Continual world: {A} robotic benchmark for continual reinforcement
  learning.
\newblock In \emph{NeurIPS}, 2021.

\bibitem[Wolczyk et~al.(2022)Wolczyk, Zajac, Pascanu, Kucinski, and
  Milos]{corr/clonex-sac}
Wolczyk, M., Zajac, M., Pascanu, R., Kucinski, L., and Milos, P.
\newblock Disentangling transfer in continual reinforcement learning.
\newblock \emph{CoRR}, 2022.

\bibitem[Wortsman et~al.(2020)Wortsman, Ramanujan, Liu, Kembhavi, Rastegari,
  Yosinski, and Farhadi]{nips/supsup}
Wortsman, M., Ramanujan, V., Liu, R., Kembhavi, A., Rastegari, M., Yosinski,
  J., and Farhadi, A.
\newblock Supermasks in superposition.
\newblock In \emph{NeurIPS}, 2020.

\bibitem[Yang et~al.(2022)Yang, Jiang, Zhou, Ma, and Shi]{iclr/p3}
Yang, Y., Jiang, J., Zhou, T., Ma, J., and Shi, Y.
\newblock Pareto policy pool for model-based offline reinforcement learning.
\newblock In \emph{ICLR}, 2022.

\bibitem[Yu et~al.(2019)Yu, Quillen, He, Julian, Hausman, Finn, and
  Levine]{corl/metaworld}
Yu, T., Quillen, D., He, Z., Julian, R., Hausman, K., Finn, C., and Levine, S.
\newblock Meta-world: {A} benchmark and evaluation for multi-task and meta
  reinforcement learning.
\newblock In \emph{CoRL}, 2019.

\bibitem[Yu et~al.(2020)Yu, Thomas, Yu, Ermon, Zou, Levine, Finn, and
  Ma]{nips/mopo}
Yu, T., Thomas, G., Yu, L., Ermon, S., Zou, J.~Y., Levine, S., Finn, C., and
  Ma, T.
\newblock {MOPO:} model-based offline policy optimization.
\newblock In \emph{NeurIPS}, 2020.

\bibitem[Zhao et~al.(2023)Zhao, Zhou, Long, Jiang, and Zhang]{icml/zhaohy}
Zhao, H., Zhou, T., Long, G., Jiang, J., and Zhang, C.
\newblock Does continual learning equally forget all parameters?
\newblock In \emph{ICML}, 2023.

\end{thebibliography}
